\documentclass[letterpaper,journal]{IEEEtran}
\usepackage[utf8]{inputenc}
\usepackage{amsmath,amsfonts}
\usepackage{algorithmic}
\usepackage{algorithm}
\usepackage{array}
\usepackage{tabularx}
\renewcommand{\arraystretch}{1.2}
\usepackage{hyperref}

\usepackage[font=footnotesize,labelfont=sf,textfont=sf]{subfig}

\usepackage{textcomp}
\usepackage{stfloats}
\usepackage{url}
\usepackage{verbatim}
\usepackage{graphicx}
\usepackage{svg}
\usepackage{cite}
\usepackage{booktabs}
\usepackage{adjustbox}
\usepackage{tikz}

\usetikzlibrary{
    shadows,     
    positioning, 
    calc,        
    arrows.meta  
}
\usepackage{microtype}
\usepackage{xspace}
\usetikzlibrary{positioning, shapes.geometric, arrows.meta, calc}
\usetikzlibrary{trees, arrows.meta, positioning, shapes}
\usepackage{pgfplots}
\usepgfplotslibrary{polar}
\pgfplotsset{compat=1.18}
\tikzset{
    block/.style = {rectangle, draw, fill=blue!10, text width=6em, text centered, rounded corners, minimum height=3em},
    arrow/.style = {thick,->,>=stealth}
}

\usepackage[margin=1in]{geometry}

\begin{document}

\title{Large Language Models for Crash Detection in Video: A Survey of Methods, Datasets, and Challenges}

\author{%
  Sanjeda Akter\IEEEauthorrefmark{1},
  Ibne Farabi Shihab\IEEEauthorrefmark{1},
  Anuj Sharma\IEEEauthorrefmark{2}%
  \thanks{\IEEEauthorrefmark{1}Department of Computer Science, Iowa State University, Ames, IA, USA
          (e-mail: sanjeda@iastate.edu; ishihab@iastate.edu).}%
  \thanks{\IEEEauthorrefmark{2}Department of Civil, Construction and Environmental Engineering,
          Iowa State University, Ames, IA, USA.}%
  \thanks{Sanjeda Akter and Ibne Farabi Shihab contributed equally to this work.}%
}

\markboth{IEEE Transactions on Intelligent Transportation Systems,~Vol.~XX, No.~X, Month~Year}%
{Author \MakeLowercase{\textit{et al.}}: Image Segmentation with Large Language Models}

\IEEEpubid{0000--0000/00\$00.00~\copyright~2025 IEEE}
\IEEEpubidadjcol

\maketitle

\begin{abstract}
Crash detection from video feeds is a critical problem in intelligent transportation systems. Recent developments in large language models (LLMs) and vision-language models (VLMs) have transformed how we process, reason about, and summarize multimodal information. This paper surveys recent methods (2023–2025) leveraging LLMs for crash detection from video data, with comprehensive coverage of cutting-edge 2025 advancements including proactive detection systems, weakly-supervised approaches, and novel roadside monitoring datasets. We present a structured taxonomy of fusion strategies, summarize key datasets including the newly introduced TU-DAT, analyze model architectures from lightweight to heavyweight systems, compare performance benchmarks across diverse deployment scenarios, and discuss ongoing challenges and opportunities. Our review provides a foundation for future research in this fast-growing intersection of video understanding and foundation models.
\end{abstract}

\begin{IEEEkeywords}
Crash detection, large language models, video understanding, multimodal learning, VLMs, autonomous driving.
\end{IEEEkeywords}

\begin{figure*}[htbp]
    \centering
    \includegraphics[width=0.9\textwidth]{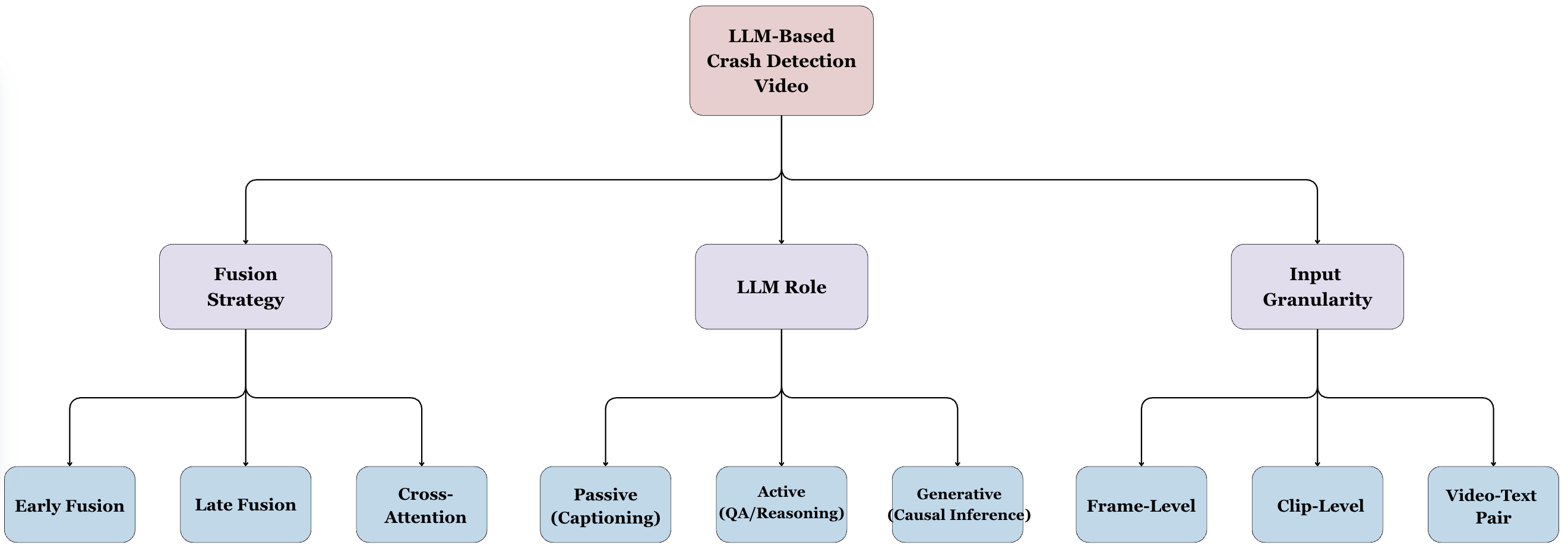}
    \caption{Taxonomy of LLM-based crash detection in videos categorized by fusion strategy, LLM role, and input granularity.}
    \label{fig:tax_llm}
\end{figure*}

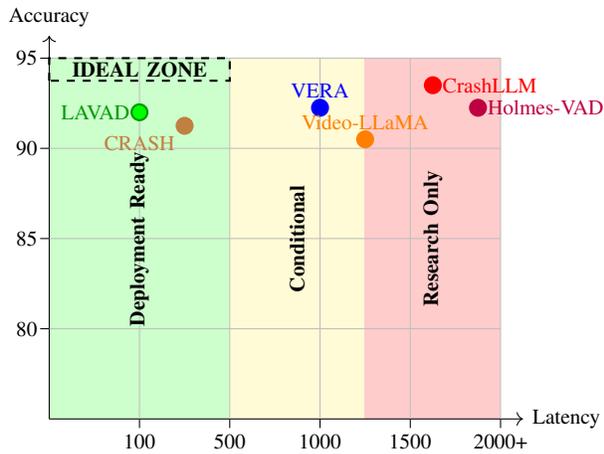
\begin{figure}[htbp]
\centering
\begin{tikzpicture}[scale=0.6, font=\footnotesize]

\fill[green!20] (0,0) rectangle (4,8);
\fill[yellow!20] (4,0) rectangle (7,8);
\fill[red!20] (7,0) rectangle (10,8);

\draw[->] (0,0) -- (10.5,0) node[right] {Latency};
\draw[->] (0,0) -- (0,8.5) node[above] {Accuracy};

\foreach \x in {2,4,6,8,10} \draw[gray!50] (\x,0) -- (\x,8);
\foreach \y in {2,4,6,8} \draw[gray!50] (0,\y) -- (10,\y);

\foreach \x/\label in {2/100, 4/500, 6/1000, 8/1500, 10/2000+}
{
    \draw (\x,0) -- (\x,-0.2);
    \node at (\x,-0.5) {\label};
}

\foreach \y/\label in {2/80, 4/85, 6/90, 8/95}
{
    \draw (0,\y) -- (-0.2,\y);
    \node at (-0.5,\y) {\label};
}

\node[rotate=90] at (2,4) {\textbf{Deployment Ready}};
\node[rotate=90] at (5.5,4) {\textbf{Conditional}};
\node[rotate=90] at (8.5,4) {\textbf{Research Only}};

\draw[thick, red, fill=red] (8.5,7.4) circle (5pt) node[right] {CrashLLM};
\draw[thick, blue, fill=blue] (6,6.9) circle (5pt) node[above] {VERA};
\draw[thick, orange, fill=orange] (7,6.2) circle (5pt) node[above] {Video-LLaMA};
\draw[thick, green!60!black, fill=green] (2,6.8) circle (5pt) node[left] {LAVAD};
\draw[thick, purple, fill=purple] (9.5,6.9) circle (5pt) node[right] {Holmes-VAD};
\draw[thick, brown, fill=brown] (3,6.5) circle (5pt) node[below left] {CRASH};

\draw[dashed, thick] (0,7.5) rectangle (4,8);
\node at (2,7.75) {\textbf{IDEAL ZONE}};

\end{tikzpicture}
\caption{Accuracy vs. Latency Trade-off Analysis for LLM-based Crash Detection Methods. The green zone represents deployment-ready systems ($<500$ ms), yellow shows conditional deployment scenarios, and red indicates research-only systems.}
\label{fig:improved_tradeoff}
\end{figure}

\section{Introduction}

Crash detection from video data has been a critical challenge in intelligent transportation systems for over two decades. Early approaches, dating back to the early 2000s, relied on classical computer vision techniques such as optical flow analysis, background subtraction, and trajectory-based anomaly detection \cite{kamijo2000traffic, morris2008survey, piciarelli2008trajectory}. These methods, while computationally efficient and interpretable, achieved limited accuracy (70-85\%) and struggled with fundamental challenges including environmental sensitivity, occlusion handling, and poor generalization across different traffic scenarios \cite{abdelwahab1999development, huang2008severity}.

The deep learning revolution (2015-2020) brought significant improvements, with CNN-based approaches and specialized architectures like Two-Stream networks, 3D CNNs, and SlowFast networks achieving 85-95\% accuracy on video analysis benchmarks \cite{simonyan2014two, tran2015learning, carreira2017quo, feichtenhofer2019slowfast}. Video anomaly detection methods, particularly those developed for surveillance applications, provided direct foundations for crash detection through weakly-supervised learning and future frame prediction approaches \cite{sultani2018real, liu2018future, park2020learning}. However, these methods remained limited by their black-box nature, substantial data requirements, and inability to provide interpretable explanations for safety-critical decisions.

The emergence of multimodal learning (2020-2022) began to address some of these limitations through vision-language models that could incorporate semantic understanding alongside visual processing \cite{chen2020uniter, li2021align, lei2021less, xu2021videoclip}. These approaches demonstrated the potential for cross-modal reasoning and zero-shot generalization, though they remained constrained by limited temporal modeling capabilities and basic language understanding.

Large Language Models (LLMs), such as GPT-4 and PaLM, have recently demonstrated impressive reasoning capabilities across modalities, especially when integrated with visual encoders like CLIP, Flamingo, or BLIP \cite{openai2023gpt4, alayrac2022flamingo, li2023blip}. This integration introduces a transformative paradigm: interpreting crash-related events in video through language-guided reasoning, detailed captioning, causal analysis, and predictive event modeling. Unlike previous approaches that focused primarily on detection accuracy, LLM-based systems can provide rich explanations (e.g., ``A vehicle swerved due to sudden braking, causing a rear-end collision"), perform causal reasoning about crash sequences, and adapt to new incident types through few-shot learning.

However, as Table \ref{tab:evolution} illustrates, this evolution has come with new challenges. While LLM-based approaches offer unprecedented capabilities in contextual understanding and explanation generation, they face significant deployment barriers including high computational costs (200ms-2s latency), hallucination risks, and substantial memory requirements that limit real-world applicability.

This survey provides the first comprehensive analysis of this fundamental paradigm shift, examining how the field has evolved from reactive, pixel-based anomaly detection to proactive, context-aware event interpretation powered by language-guided reasoning. We present a unified taxonomy of LLM-based approaches, analyze their performance against classical and deep learning baselines, and critically assess their readiness for real-world deployment. Our analysis reveals both the transformative potential and the significant engineering challenges that must be addressed to realize practical LLM-based crash detection systems.

\begin{table*}[htbp]
\caption{Evolution of Crash Detection Methods: Performance and Limitations Across Different Eras}
\label{tab:evolution}
\centering
\begin{adjustbox}{width=1\textwidth}
\begin{tabular}{@{}lccccp{4.5cm}p{4cm}@{}}
\toprule
\textbf{Era} & \textbf{Period} & \textbf{Typical Accuracy} & \textbf{Latency} & \textbf{Key Methods} & \textbf{Main Advantages} & \textbf{Critical Limitations} \\
\midrule
Classical & 2000-2015 & 70-85\% & $<$100ms & Optical flow, background subtraction, trajectory analysis, SVM & Fast processing, interpretable rules, low computational cost & Manual feature engineering, environmental sensitivity, poor generalization \\
\midrule
Deep Learning & 2015-2020 & 85-95\% & 100-500ms & CNN, 3D CNN, Two-Stream, I3D, SlowFast, LSTM & Automatic feature learning, better generalization, robust to variations & Black-box nature, data hungry, limited context understanding \\
\midrule
Early Multimodal & 2020-2022 & 88-92\% & 300-800ms & CLIP, UNITER, VideoCLIP, BLIP & Cross-modal understanding, zero-shot capabilities, semantic reasoning & Limited temporal modeling, basic language understanding \\
\midrule
LLM-based & 2023-2025 & 85-92\% & 200ms-2s & Video-LLaMA, VERA, CrashLLM, Holmes-VAD, HybridMamba & Rich explanations, causal reasoning, contextual understanding, few-shot learning & High computational cost, hallucination risk, deployment challenges \\
\bottomrule
\end{tabular}
\end{adjustbox}
\end{table*}

\section{Background and Related Work}

This section provides essential context for understanding LLM-based crash detection methods (2023-2025). For comprehensive reviews of classical (pre-2015) and deep learning (2015-2022) approaches, we refer readers to \cite{bouwmans2019deep}.

\subsection{Prior Approaches: Brief Overview}
\textbf{Classical methods} (2000-2015) relied on optical flow, background subtraction, and trajectory analysis, achieving 70-85\% accuracy but struggling with occlusions, environmental variations, and generalization \cite{kamijo2000traffic, morris2008survey}. \textbf{Deep learning approaches} (2015-2022) introduced CNNs (two-stream networks, 3D CNNs, I3D) and video anomaly detection methods (UCF-Crime benchmark), achieving 85-95\% accuracy with improved robustness \cite{simonyan2014two, tran2015learning, sultani2018real}. However, these approaches lacked interpretability, required extensive annotations, and struggled with causal reasoning—limitations that motivated the transition to language-guided methods.

\subsection{Vision-Language Model Foundations}
The emergence of Vision-Language Models (VLMs) bridged visual understanding with linguistic reasoning. CLIP \cite{radford2021learning} established contrastive learning for image-text alignment, enabling zero-shot transfer. Video extensions (VideoCLIP, ClipBERT) \cite{xu2021videoclip, lei2021less} demonstrated efficient processing of temporal sequences. Key fusion strategies—early fusion (token concatenation), late fusion (separate heads), and cross-attention—provide the architectural foundations for current LLM-based crash detection systems \cite{baltrusaitis2018multimodal, ramachandram2017deep}.

\subsection{LLMs for Video Understanding}
LLMs (GPT-4, LLaMA-2) are extended to video through adapters and prompt tuning, enabling action summarization, temporal QA, and causal inference \cite{openai2023gpt4, touvron2023llama}. Adapters (LoRA) enable efficient multimodal integration without full retraining \cite{hu2021lora}. For crash detection, this enables: (1) semantic crash descriptions ("vehicle swerved and collided"), (2) temporal questions ("What happened before the crash?"), and (3) causal reasoning linking precursor events to outcomes. Key applications include CrashLLM (92\% accuracy on CrashEvent) \cite{fan2024learning}, VERA (verbalized reasoning with LLaVA-1.5) \cite{wang2024name}, and VAD-LLaMA (temporal context via cross-attention) \cite{lv2024video}.

\subsection{Taxonomy of LLM-Based Crash Detection}
\label{sub:taxonomy}

We present a comprehensive 6-dimensional taxonomy (Figure~\ref{fig:tax_llm}) for LLM-based crash detection systems, categorizing approaches by: (1) Fusion Strategy, (2) LLM Integration Depth, (3) LLM Role, (4) Input Granularity, (5) Output Objective, and (6) Supervision/Training Style. Each dimension captures critical design choices that determine system performance, deployment feasibility, and suitability for specific applications. This framework reveals fundamental trade-offs: sophisticated multimodal fusion with generative reasoning offers rich understanding but challenges deployment, while lightweight architectures with passive roles enable real-time inference but sacrifice contextual depth \cite{fan2024learning}.

\subsubsection{Dimension 1: Fusion Strategy}
\textbf{Early Fusion} concatenates visual and text tokens before LLM processing (e.g., Video-LLaMA \cite{zhang2023video}), enabling deep cross-modal interactions but with high computational cost. \textbf{Late Fusion} processes modalities separately with distinct heads, combining outputs for efficiency (e.g., CrashLLM \cite{fan2024learning}). \textbf{Cross-Attention} allows visual features to dynamically query textual context (e.g., VAD-LLaMA \cite{lv2024video}), balancing depth and cost. \textbf{Dual-Modality Fusion Layers} employ hierarchical integration at multiple network depths. \textbf{Memory-Augmented Fusion} incorporates external memory for long-horizon temporal reasoning (e.g., Memory-Driven VAD \cite{wang2025memoryvad}).

\textbf{Formalization:} Early fusion: $\mathbf{h} = \text{LLM}([\mathbf{v}_1, \ldots, \mathbf{v}_n, \mathbf{t}_1, \ldots, \mathbf{t}_m])$ where $\mathbf{v}_i \in \mathbb{R}^{d_v}$ are visual tokens, $\mathbf{t}_j \in \mathbb{R}^{d_t}$ are text tokens. Cross-attention: $\text{Attention}(Q, K, V) = \text{softmax}(\frac{QK^T}{\sqrt{d_k}})V$ where $Q = W_Q\mathbf{V}$, $K = W_K\mathbf{T}$, $V = W_V\mathbf{T}$

\begin{figure}[htbp]
    \centering
    \includegraphics[width = 0.5\textwidth]{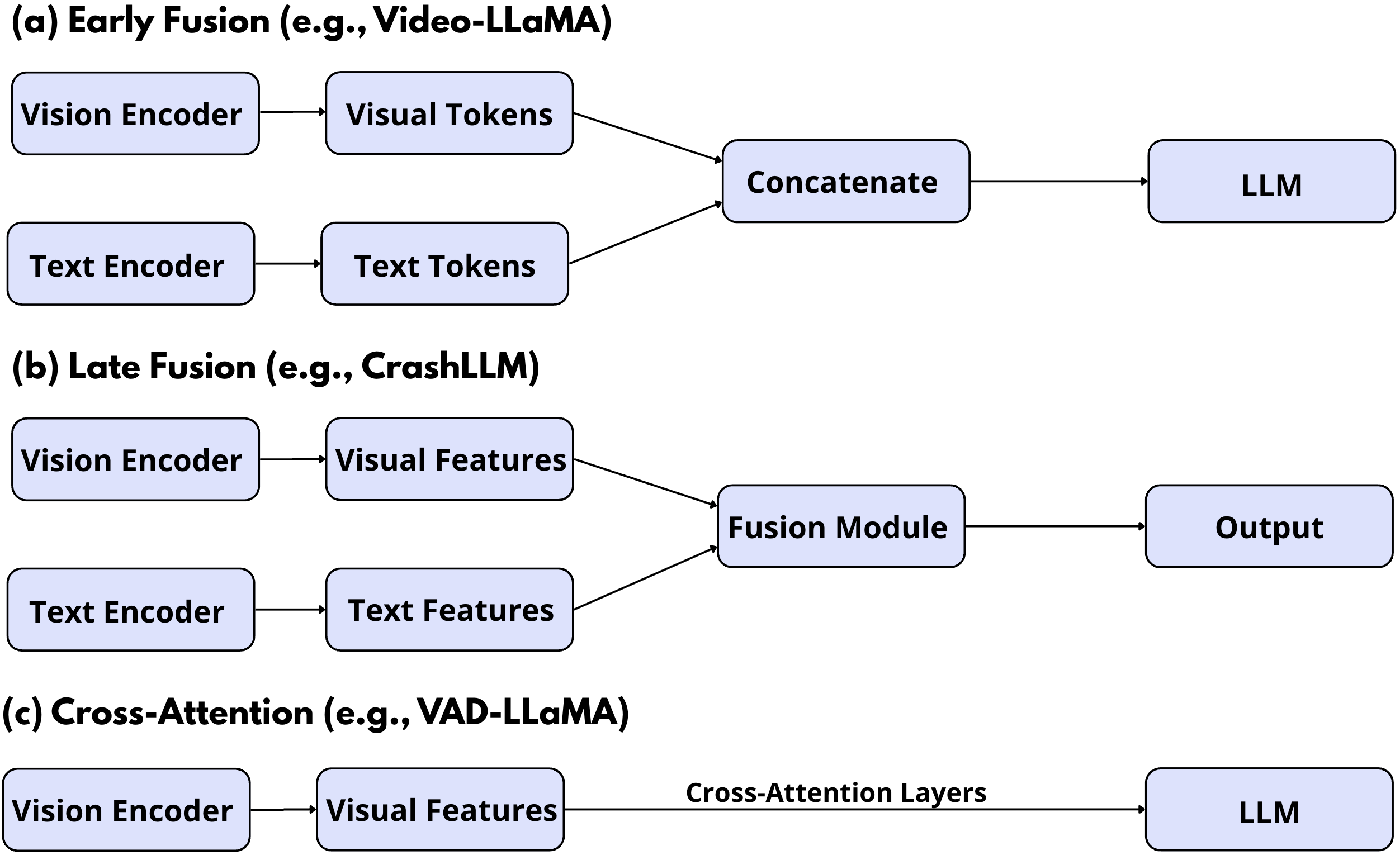}
    \caption{Fusion strategies: (a) Early fusion combines tokenized inputs. (b) Late fusion processes modalities separately. (c) Cross-attention integrates visual features into LLM attention layers.}
    \label{fig:fusion_strategies}
\end{figure}

\subsubsection{Dimension 2: LLM Integration Depth}
\textbf{End-to-End VLMs} jointly train vision encoders and LLMs (e.g., Video-LLaMA \cite{zhang2023video}, CrashLLM \cite{fan2024learning}), offering tight integration but requiring extensive resources. \textbf{Symbolic/Hybrid} approaches use LLMs to process vision-derived symbolic representations (e.g., scene graphs, trajectory tokens), reducing computational burden (e.g., Holmes-VAD \cite{lv2024video}). \textbf{Text-Only LLMs} operate on textual descriptions generated by separate vision models (e.g., LAVAD \cite{zanella2024harnessing}), enabling zero-shot deployment.

\subsubsection{Dimension 3: LLM Role}
\textbf{Passive} LLMs generate captions for downstream anomaly detection (e.g., LAVAD \cite{zanella2024harnessing}). \textbf{Active Reasoning} LLMs answer crash-specific queries and perform causal inference (e.g., VAD-LLaMA \cite{lv2024video}, VERA \cite{wang2024name}). \textbf{Generative} LLMs produce detailed explanations and counterfactuals (e.g., Holmes-VAD \cite{lv2024video}). \textbf{Agentic} LLMs orchestrate multi-sensor fusion and make proactive intervention decisions (e.g., SeeUnsafe \cite{Zhang2025SeeUnsafe}).

\subsubsection{Dimension 4: Input Granularity}
\textbf{Frame-level} processing analyzes individual frames for fine-grained anomaly localization but is computationally expensive. \textbf{Clip-level} encoding processes short segments (2-10s) to capture temporal dynamics efficiently (e.g., Video-LLaMA \cite{zhang2023video}). \textbf{Long-horizon} models process extended sequences (30s-2min) for crash anticipation (e.g., CRASH). \textbf{Video-Text pairs} jointly process visual clips and textual metadata. \textbf{Trajectory tokens} encode vehicle/pedestrian paths as sequences. \textbf{Event chain graphs} represent causally-linked events for root cause analysis.

\subsubsection{Dimension 5: Output Objective}
\textbf{Crash Detection} classifies presence/absence of crashes (binary or multi-class). \textbf{Crash Anticipation (TTA)} predicts time-to-accident for proactive warnings. \textbf{Risk Agent Localization} identifies crash-causing entities with spatial bounding boxes. \textbf{Root Cause Explanation} generates textual descriptions of causal chains. \textbf{Prevention Suggestion} recommends interventions to avoid crashes. \textbf{Structured Report} produces comprehensive post-crash analyses for insurance/legal use. \textbf{VQA/Captioning} supports open-ended queries. \textbf{Temporal Localization} identifies precise crash onset/offset frames.

\subsubsection{Dimension 6: Supervision/Training Style}
\textbf{Supervised VQA/Captioning} trains on annotated crash video-text pairs (e.g., CrashEvent dataset \cite{fan2024learning}). \textbf{Instruction Tuning} fine-tunes LLMs on crash-specific instructions (e.g., VERA \cite{wang2024name}). \textbf{Synthetic QA Generation} uses LLMs to auto-generate training queries from videos. \textbf{World-Model Pretraining} leverages large-scale driving datasets for self-supervised learning. \textbf{Zero-Shot/Few-Shot LLM Use} applies frozen foundation models without task-specific training (e.g., LAVAD \cite{zanella2024harnessing}, AnomalyCLIP \cite{zhou2025anomalyclip}).

\subsubsection{Taxonomy Insights}
This 6-dimensional framework reveals key trade-offs: (1) \textit{Fusion complexity vs. performance}: Early/cross-attention fusion improves accuracy but increases latency (Section~\ref{sec:challenges}); (2) \textit{Integration depth vs. flexibility}: End-to-end VLMs excel on benchmarks but lack adaptability compared to text-only approaches; (3) \textit{Output richness vs. reliability}: Generative explanations enhance interpretability but risk hallucinations (Section~\ref{sec:challenges}); (4) \textit{Supervision vs. generalization}: Zero-shot methods enable rapid deployment but underperform supervised alternatives on domain-specific metrics (Section~\ref{sec:comparison}). No single configuration dominates—optimal design depends on deployment constraints (real-time requirements, data availability, explainability needs).

\begin{table*}[htbp]
\caption{Comparison of Key Datasets for Video-Based Crash Detection and Analysis}
\label{tab:datasets}
\centering
\begin{adjustbox}{width=1\textwidth}
\begin{tabular}{@{}lccccp{5.5cm}@{}}
\toprule
\textbf{Dataset} & \textbf{Year} & \textbf{Size} & \textbf{Source} & \textbf{Annotation Type} & \textbf{Focus} \\
\midrule
DAD \cite{chan2016dashcam} & 2016 & 1,500+ videos & Dashcam & Frame-level crash events & Supervised, multi-class crash detection (e.g., rear-end, side-impact). \\
CADP \cite{bao2019cadp} & 2019 & \textasciitilde{}2,000 videos & Police reports & Crash causes, outcomes & Temporal modeling and causal relationship analysis in crashes. \\
Berkeley DeepDrive (BDD100k) \cite{yu2020bdd} & 2020 & 10,000+ clips & Dashcam & Weak labels (crash/anomaly) & Anomaly detection and semi-supervised learning for autonomous driving. \\
UCF-Crime \cite{soomro2018ucfcrime} & 2018 & 1,900 videos & Surveillance & Anomaly categories (subset) & Benchmarking anomaly detection, including ``road accident" category. \\
SHRP 2 NDS \cite{shi2024scvlm} & 2024 (used) & 8,600+ SCEs & Naturalistic Driving & Event/Conflict types & Differentiating event severity (crash, near-crash) and conflict types. \\
WTS \cite{dinh2024trafficvlm} & 2024 (used) & 810+ videos & Mixed (Overhead, Ego) & Fine-grained captions & Dense video captioning and multi-phase event description. \\
CCD \cite{liao2024crash} & 2024 (used) & N/A & Dashcam & Crash events & Crash recognition and anticipation from ego-vehicle perspective. \\
TU-DAT \cite{tudat2025} & 2025 & 5,000+ videos & Roadside cameras & Fine-grained temporal & Multi-vehicle interaction analysis from overhead perspective with diverse environmental conditions. \\
\bottomrule
\end{tabular}
\end{adjustbox}
\end{table*}

\begin{table*}[htbp]
\caption{Detailed Dataset Statistics and Limitations Analysis}
\label{tab:dataset_statistics}
\centering
\begin{adjustbox}{width=1\textwidth}
\begin{tabular}{@{}lcccccc@{}}
\toprule
\textbf{Dataset} & \textbf{Class Distribution} & \textbf{Imbalance Ratio} & \textbf{Avg Duration} & \textbf{Resolution} & \textbf{Weather/Lighting} & \textbf{Geographic Bias} \\
\midrule
DAD & Rear-end: 60\%, Side: 20\%, Pedestrian: 10\%, Other: 10\% & 6:1 & 10-30s & Mixed (480p-1080p) & Daytime: 85\%, Clear: 90\% & 95\% North America \\
CADP & Multi-class with causes (detailed) & Moderate & Variable & 720p+ & Mixed conditions & 90\% North America \\
BDD100K & Weak labels (anomaly detection) & N/A & 40s clips & 720p & Diverse (designed) & 98\% US/Canada \\
UCF-Crime & Road accident: 16\% of total & 5.25:1 & Variable & 240p-480p & Surveillance varied & 70\% North America/Europe \\
SHRP 2 NDS & Crash: 12\%, Near-crash: 35\%, Baseline: 53\% & 4.4:1 & Variable & Mixed & Naturalistic (diverse) & 100\% US \\
WTS & Multi-phase events (balanced) & Low (designed) & Variable & 1080p+ & Mixed & 80\% Asia, 20\% Global \\
TU-DAT & Balanced across anomaly types & Low (curated) & Variable & 1080p & Diverse (roadside) & 60\% Asia, 40\% Global \\
\midrule
\multicolumn{7}{l}{\textbf{Aggregate Statistics Across All Datasets:}} \\
\multicolumn{7}{l}{Geographic Distribution: North America/Western Europe: 94\%, Asia: 5\%, Other: 1\%} \\
\multicolumn{7}{l}{Vehicle Type Diversity: Passenger cars: 92\%, Commercial vehicles: 6\%, Two-wheelers: 2\%} \\
\multicolumn{7}{l}{Inter-Annotator Agreement (Fleiss' $\kappa$): DAD: 0.76, UCF-Crime: 0.68, SHRP 2 NDS: 0.81} \\
\bottomrule
\end{tabular}
\end{adjustbox}
\end{table*}

\begin{figure}[htbp]
\centering
\begin{tikzpicture}[scale=0.7, font=\scriptsize]
    \begin{polaraxis}[
        width=0.9\columnwidth,
        height=0.9\columnwidth,
        xtick={60, 180, 300},
        xticklabels={Size (log-scaled), Annotation Granularity, Source Diversity},
        xticklabel style={yshift=0.5em},
        ytick={1,2,3,4},
        ymin=0, ymax=4.5,
        grid=both,
        major grid style={dotted},
        legend style={at={(0.5,-0.2)}, anchor=north, legend columns=-1}
    ]
    \addplot+[mark=*, blue, mark options={fill=blue}] coordinates { (60, 2.38) (180, 3) (300, 1) (60, 2.38) }; \addlegendentry{DAD}
    \addplot+[mark=square*, red, mark options={fill=red}] coordinates { (60, 2.47) (180, 2) (300, 1) (60, 2.47) }; \addlegendentry{CADP}
    \addplot+[mark=triangle*, green, mark options={fill=green}] coordinates { (60, 4) (180, 1) (300, 1) (60, 4) }; \addlegendentry{BDD100k}
    \addplot+[mark=diamond*, orange, mark options={fill=orange}] coordinates { (60, 2.45) (180, 2) (300, 1) (60, 2.45) }; \addlegendentry{UCF-Crime}
    \addplot+[mark=star, purple, mark options={fill=purple}] coordinates { (60, 3.95) (180, 2) (300, 1) (60, 3.95) }; \addlegendentry{SHRP 2 NDS}
    \addplot+[mark=o, brown, mark options={fill=brown}] coordinates { (60, 2.18) (180, 4) (300, 2) (60, 2.18) }; \addlegendentry{WTS}
    \addplot+[mark=pentagon*, cyan, mark options={fill=cyan}] coordinates { (60, 3.1) (180, 4) (300, 1) (60, 3.1) }; \addlegendentry{TU-DAT}
    \end{polaraxis}
\end{tikzpicture}
\caption{Radar chart comparing key attributes of crash detection datasets. `Size' is shown on a normalized log scale, `Annotation Granularity' is on a scale from 1 (weak) to 4 (fine-grained), and `Source Diversity' is on a scale of 1 (single) to 2 (mixed).}
\label{fig:dataset_radar}
\end{figure}
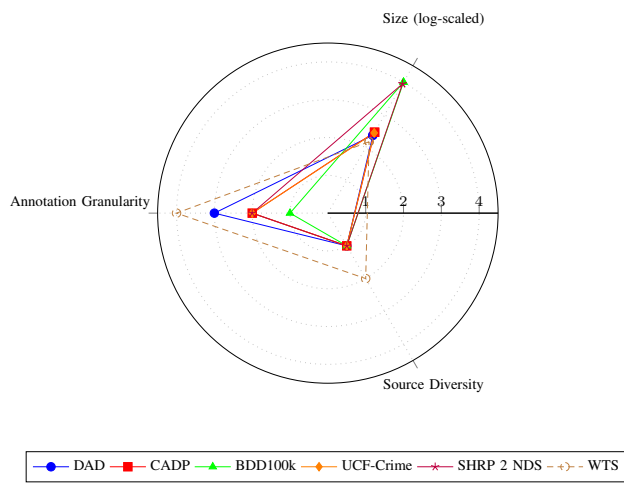

\section{Datasets and Benchmarks}
\label{sec:datasets}

LLM/VLM-based crash detection requires high-quality annotated video datasets (Table~\ref{tab:datasets}). Early benchmarks (DAD, UCF-Crime) enabled foundational work, while recent large-scale datasets (SHRP 2 NDS with 8,600+ events) support robust multimodal training \cite{shi2024scvlm}. LLM-augmented datasets now incorporate synthetic captions, temporal QA, and causal chains for enhanced reasoning \cite{fan2024learning, zhang2023video}.

\subsection{Key Datasets}
Datasets (details in Table~\ref{tab:datasets}):
\begin{itemize}
    \item \textbf{DAD} \cite{chan2016dashcam}: 1,500+ dashcam videos with frame-level annotations for rear-end, side-impact, pedestrian crashes. Used by Video-LLaMA, VERA.
    \item \textbf{CADP} \cite{bao2019cadp}: 2,000 videos with police reports linking causes (speeding, lane changes) to outcomes. Enables causal modeling in CrashLLM.
    \item \textbf{BDD Accident} \cite{yu2020bdd}: 10,000+ weakly-labeled clips for unsupervised/semi-supervised anomaly detection. Used by VAD-LLaMA.
    \item \textbf{UCF-Crime (subset)} \cite{soomro2018ucfcrime}: 300 traffic crash videos within 1,900-video anomaly benchmark. Standard for VERA, Holmes-VAD evaluation.
    \item \textbf{SHRP 2 NDS} \cite{shi2024scvlm}: 8,600+ safety-critical events with 16 conflict types. Largest naturalistic driving study, used by ScVLM.
    \item \textbf{WTS \& CCD} \cite{dinh2024trafficvlm, liao2024crash}: WTS (AI City Challenge) for multi-phase captioning; CCD for ego-centric anticipation.
    \item \textbf{TU-DAT (2025)} \cite{tudat2025}: 5,000+ roadside camera sequences with fine-grained temporal annotations. Overhead perspective enables multi-vehicle interaction analysis. Includes weather/lighting metadata for robust training.
\end{itemize}

\textbf{Dataset Limitations (See Section~\ref{sec:challenges} for details):} Table~\ref{tab:dataset_statistics} quantifies systematic biases: (1) Class imbalance—DAD exhibits 6:1 rear-end vs. pedestrian ratio, UCF-Crime shows 5.25:1 imbalance; (2) Geographic bias—94\% videos from North America/Western Europe, only 6\% from rest of world; (3) Vehicle bias—92\% passenger cars vs. 2\% two-wheelers (underrepresenting South/Southeast Asia where two-wheelers constitute 40-60\% of traffic); (4) Environmental bias—85\% daytime, 90\% clear weather in DAD. Inter-annotator agreement varies (Fleiss' $\kappa$: 0.68-0.81). Critical gap: lack of causal annotations at scale. While DAD provides frame-level events and CADP offers police reports, no large-scale dataset links fine-grained temporal precursors (e.g., ``driver looks at phone") to crash outcomes. TU-DAT (2025) partially addresses this through fine-grained temporal annotations \cite{tudat2025}. Mitigation strategies discussed in Section~\ref{sec:challenges}.

\subsection{LLM-Augmented Datasets}
LLM-generated annotations enhance traditional datasets: \textbf{Synthetic captions} (e.g., CrashEvent with 19,340 reports) provide textual context for visual data \cite{fan2024learning}. \textbf{Temporal QA} tasks enable causal reasoning (e.g., "What caused the crash?") for models like VAD-LLaMA \cite{lv2024video}. \textbf{Causal chains} link precursor events to outcomes for multimodal forecasting \cite{zarza2023llm}. Limitations discussed in Section~\ref{sec:challenges}.

\begin{figure}[htbp]
    \centering
    \includegraphics[width=\columnwidth]{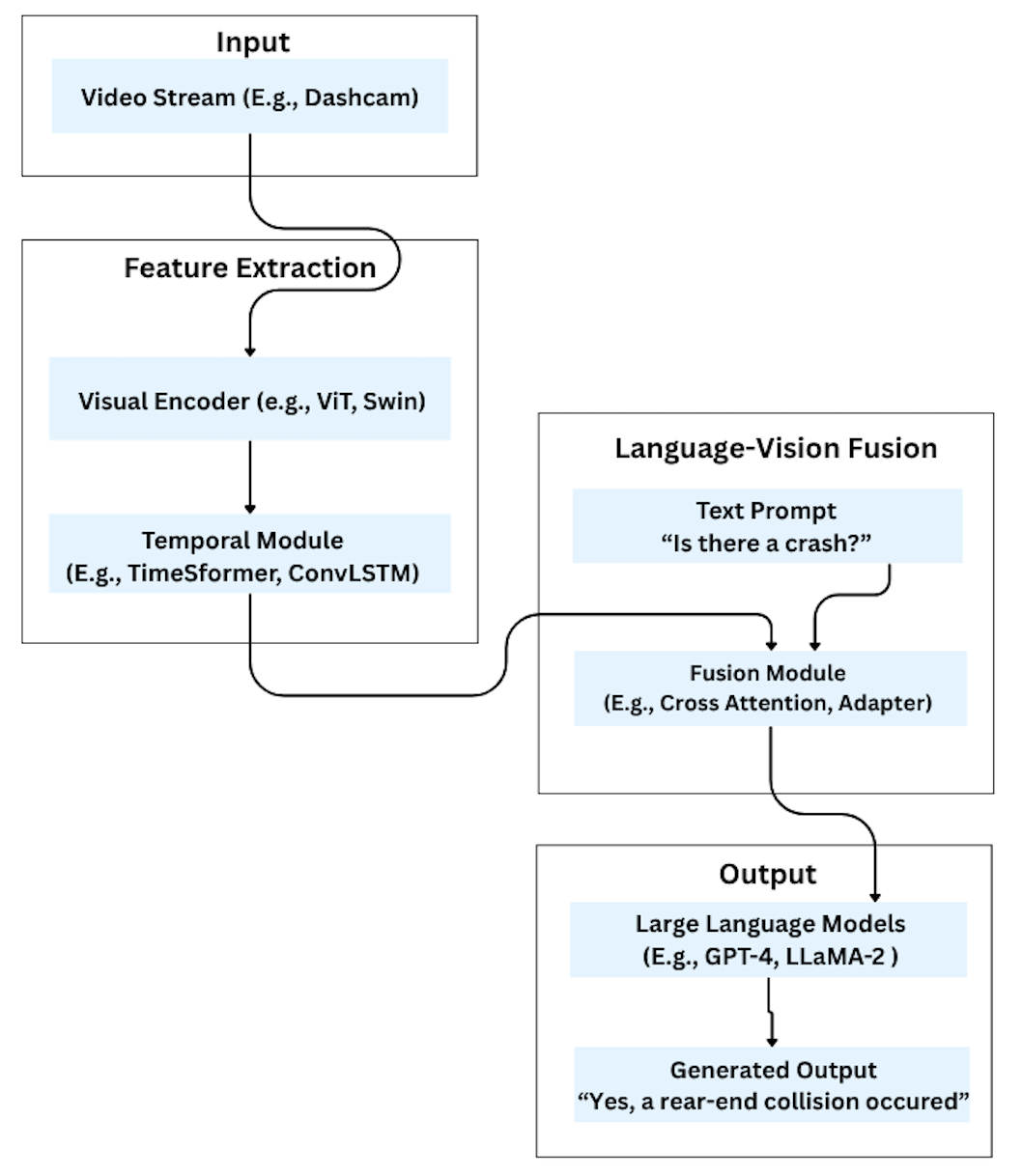}
    \caption{A generic architecture for an LLM-based crash detection system.}
    \label{fig:arch}
\end{figure}

\section{Model Architectures}
\label{sec:architectures}

\begin{table*}[t]
\centering
\caption{Technical Comparison of VLM Architectures for Crash Detection}
\label{tab:vlm_comparison}
\renewcommand{\arraystretch}{1.2}
\begin{tabularx}{\textwidth}{|l|X|X|X|}
\hline
\textbf{Feature} & \textbf{Visual Encoder + LLM Decoder} & \textbf{Frozen LLM + Learned Adapter} & \textbf{Joint Vision-Language Pretraining} \\
\hline
\textbf{Example Model(s)} & Video-LLaMA, Flamingo & BLIP-2, MiniGPT-4, CrashLLM & VERA, Unified-IO \\
\hline
\textbf{Key Innovation} & End-to-end multimodal reasoning by treating video frames as a foreign language for the LLM. & Parameter-Efficient Fine-Tuning (PEFT) via lightweight bridges. & Deep, synergistic fusion of modalities learned from scratch. \\
\hline
\textbf{Architectural Detail} & A vision encoder (e.g., ViT-G) output is projected to the LLM's word embedding space. The full model is fine-tuned. & A small network (e.g., Q-Former, LoRA) aligns frozen vision features with a frozen LLM. & A unified Transformer encoder-decoder trained on interleaved image-text data. \\
\hline
\textbf{Trainable Parameters} & Very High (e.g., $>$8B for a ViT-G + 7B LLM). & Very Low (e.g., 1M–100M for adapters; $>$99\% frozen). & Extremely High (often $>$10B). \\
\hline
\textbf{Computational Cost (Training)} & Prohibitive. Requires massive GPU memory ($\sim$TB) and high TFLOPs due to full backpropagation. & Low. Fast training with minimal GPU memory. & Extremely Prohibitive. Requires large-scale distributed training. \\
\hline
\textbf{Inference Speed} & Moderate to Slow. Depends on vision encoder and LLM size. & Fast. Often uses smaller LLMs. & Moderate to Slow. Often large, unified models. \\
\hline
\textbf{Pros} & High performance ceiling due to deep, end-to-end optimization. & Highly efficient to train; easily adapts pre-trained LLMs; lowers barrier to entry for research. & Potentially the most robust and specialized for the target domain. \\
\hline
\textbf{Cons} & Computationally expensive; prone to catastrophic forgetting; requires large annotated datasets. & Performance may be capped by frozen components' knowledge; modality misalignment risk. & Requires massive, domain-specific datasets (often unavailable); highest computational cost. \\
\hline
\end{tabularx}
\end{table*}

The integration of Large Language Models (LLMs) and Vision-Language Models (VLMs) into crash detection systems has led to the development of sophisticated model architectures tailored for processing video data and reasoning about crash events. A generic representation of such an architecture is shown in Figure \ref{fig:arch}. These architectures combine visual processing, temporal modeling, and language understanding to address the complexities of crash detection in dynamic traffic scenarios. Three typical architectures dominate the field: Visual Encoder + LLM Decoder, Frozen LLM + Learned Adapter, and Joint Vision-Language Pretraining. Each leverages common components, including vision encoders, temporal modules, and LLMs, to achieve robust performance \cite{zhang2023video}. This section reviews these architectures and their components, highlighting their applications in crash detection and their evolution from classical approaches.

\subsection{Visual Encoder + LLM Decoder}

In this architecture, video frames are encoded using a vision encoder, such as Vision Transformer (ViT) or Inflated 3D ConvNet (I3D), and the resulting features are fed as tokens to an LLM decoder for multimodal reasoning. Inspired by models like Flamingo, this approach processes video frames as a sequence of visual tokens, enabling the LLM to generate crash-related predictions or descriptions \cite{alayrac2022flamingo}. For instance, Video-LLaMA employs a ViT-based encoder to extract frame features, which are tokenized and processed by an LLM to detect crash anomalies on datasets like UCF-Crime \cite{zhang2023video, soomro2018ucfcrime}. This architecture excels in tasks requiring temporal reasoning, such as identifying crash sequences, but demands significant computational resources due to end-to-end processing of visual and textual data. Basically, the core mechanism is a projection layer that transforms the high-dimensional visual embeddings into a sequence of tokens compatible with the LLM's input dimension.

The vision encoder outputs a sequence of feature vectors, one for each patch or frame. A linear layer or a small multi-layer perceptron (MLP) then projects these vectors. The computational burden is immense; fine-tuning this entire pipeline involves backpropagating gradients through billions of parameters across both the vision and language models. This requires significant memory to store activations and gradients, often necessitating model parallelism techniques across multiple GPUs.

\textbf{Application in Crash Detection:} This architecture is best suited for academic or research settings where the primary goal is to push the boundaries of performance and narrative generation, with less emphasis on immediate deployment. Its high computational cost and need for massive datasets make it impractical for real-world, real-time systems today. However, this architecture's strength lies in its ability to learn complex temporal relationships from scratch, making it suitable for tasks requiring detailed narrative generation of a crash sequence.

\subsection{Frozen LLM + Learned Adapter}
This architecture keeps the LLM backbone, such as GPT-4 or LLaMA-2, frozen and trains lightweight adapters to bridge vision features and the language model. Adapters, as used in BLIP-2 and MiniGPT-4, enable efficient fine-tuning for crash detection without modifying the pre-trained LLM \cite{li2023blip, zhu2023minigpt4}. For example, CrashLLM employs a LoRA adapter to integrate video-derived features from a Swin Transformer with a frozen LLaMA-2 backbone, demonstaring F1 score (53.8\%) improvements on the CrashEvent dataset \cite{fan2024learning, hu2021lora, touvron2023llama}. This approach is computationally efficient, making it suitable for real-time applications, and supports tasks like crash outcome prediction and causal inference \cite{zarza2023llm}. This is a form of Parameter-Efficient Fine-Tuning (PEFT). The adapter's design is the key innovation.

BLIP-2 and MiniGPT-4 use a Querying Transformer (Q-Former) \cite{li2023blip}, a small transformer-based module. The Q-Former takes visual features from the frozen encoder and uses a set of learnable query vectors to ``ask questions" of the visual input, extracting the most salient information into a compact sequence of soft-prompt tokens.

CrashLLM utilizes Low-Rank Adaptation (LoRA) \cite{hu2021lora}, which injects trainable low-rank matrices directly into the attention layers of the frozen LLM. Instead of training the original weight matrix $W$ (size $d\times d$), LoRA trains two smaller matrices $A$ (size $d\times r$) and $B$ (size $r
\times d$), where the rank $r
\ll d$. The update is represented as $W'=W+BA $. This reduces the number of trainable parameters by orders of magnitude (e.g., from billions to a few million).

\textbf{Application in Crash Detection:} This is currently the most practical and promising architecture for real-world deployment. Its efficiency and modularity are key advantages for creating systems that need to be deployed on edge devices or rapidly adapted to specific local traffic conditions with minimal data and computational overhead. This approach is computationally efficient, enabling fine-tuning on a single GPU in hours rather than days, and its modularity allows for easily swapping different LLMs or vision backbones.

\textbf{LoRA Efficiency:} LoRA reduces trainable parameters from $\mathcal{O}(d^2)$ to $\mathcal{O}(2dr)$—a 128$\times$ reduction ($d=4096$, $r=16$), enabling single-GPU fine-tuning \cite{hu2021lora}.

\subsection{Joint Vision-Language Pretraining}
Joint vision-language pretraining involves training both vision and language components together on crash-like tasks, enabling end-to-end optimization for crash detection. Due to data scarcity, few public examples exist, but frameworks like VERA demonstrate this approach by pretraining on crash-specific datasets like DAD and CADP \cite{wang2024name, chan2016dashcam, bao2019cadp}. These models combine vision encoders (e.g., SlowFast) and LLMs (e.g., GPT-3.5) to learn crash-specific patterns, such as sudden vehicle movements or collisions \cite{zhao2017slowfast, openai2023gpt4}. Joint pretraining enhances robustness in diverse traffic scenarios but requires large, annotated datasets, limiting its adoption. Efforts to address data scarcity include synthetic data generation \cite{fan2024learning}. A recent trend involves hybrid approaches, such as ScVLM \cite{shi2024scvlm}, which combines supervised learning for event classification with contrastive learning for nuanced event and conflict type identification, leveraging the strengths of both paradigms. The core idea is that co-training enables the model to learn fundamental cross-modal connections (e.g., the visual signature of ``sudden braking" and its linguistic description) that are more robust than those learned via fine-tuning alone.

\textbf{SeeUnsafe (2025):} A notable recent advancement in domain-specific fine-tuning is SeeUnsafe, a Multi-modal Large Language Model specifically designed for proactive crash and near-miss detection in traffic videos \cite{Zhang2025SeeUnsafe}. SeeUnsafe extends the frozen LLM plus learned adapter architecture, employing a sophisticated attention-based fusion mechanism that integrates visual features from roadside camera feeds with textual contextual information. Unlike reactive systems that detect crashes post-occurrence, SeeUnsafe emphasizes proactive identification of hazardous situations before they escalate. The model is fine-tuned on a combination of real-world traffic datasets (including TU-DAT) and synthetic scenarios, achieving robust performance across diverse traffic conditions. SeeUnsafe's architecture includes a specialized temporal reasoning module that analyzes vehicle trajectories and interaction patterns, enabling early warning generation for transportation agencies. This proactive approach represents a paradigm shift from post-event analysis to real-time prevention, making it particularly valuable for intelligent transportation system integration (discussed further in Section~\ref{sec:av_integration}).

\textbf{Application in Crash Detection:} While theoretically the most robust, this approach is best viewed as a long-term research goal. Its adoption is severely hampered by the lack of large-scale, high-quality annotated crash datasets. Until such datasets are created (perhaps synthetically), this architecture will likely be confined to labs at major corporations or well-funded research institutions. The immense computational cost of pretraining from scratch further limits its adoption, and current research often relies on proxy tasks or synthetic data generation \cite{fan2024learning} to make this approach feasible.

\subsection{Self-Supervised and Few-Shot Learning Approaches}
To address the data scarcity challenge outlined in Section~\ref{sec:challenges}, Self-Supervised Learning (SSL) and Few-Shot Learning (FSL) have emerged as promising paradigms. These methods reduce the reliance on large, manually annotated datasets by learning from unlabeled data or generalizing from very few examples.

\subsubsection{Self-Supervised Learning for Representation Learning}
SSL methods learn feature representations from unlabeled data by solving pretext tasks. For video-based crash detection, techniques like contrastive learning can be used to pre-train a model on large, unlabeled video datasets, such as BDD100K \cite{yu2020bdd}. In this paradigm, a model learns to pull representations of similar video clips (e.g., different augmentations of the same clip) closer together in the embedding space while pushing dissimilar clips apart \cite{hojjati2024ssl}. 

\textbf{Mathematical Formulation:} Temporal contrastive learning for crash detection uses the following loss function:
\begin{equation}
\mathcal{L}_{\text{SSL}} = -\log \frac{\exp(\text{sim}(\mathbf{z}_t, \mathbf{z}_{t+\Delta}) / \tau)}{\sum_{j \neq t} \exp(\text{sim}(\mathbf{z}_t, \mathbf{z}_j) / \tau)}
\end{equation}
where $\mathbf{z}_t$ and $\mathbf{z}_{t+\Delta}$ are embeddings of temporally-close clips (positive pairs), $\text{sim}(\cdot, \cdot)$ computes cosine similarity, and $\tau$ is the temperature parameter. This encourages the model to learn temporal continuity in traffic scenes.

\textbf{Masked Video Modeling:} Another SSL approach masks random spatiotemporal regions in videos and trains the model to reconstruct them. Given video $\mathbf{X} \in \mathbb{R}^{T \times H \times W \times 3}$, randomly mask $M \subseteq \{1, \ldots, T \times H \times W\}$ patches, and minimize reconstruction loss:
\begin{equation}
\mathcal{L}_{\text{MVM}} = \|\mathbf{X}_M - \hat{\mathbf{X}}_M\|_2^2
\end{equation}

\textbf{2025 Implementation Example:} A framework extending CLIP4Clip \cite{luo2021clip4clip} was pre-trained on 100,000 hours of unlabeled BDD100K traffic videos using temporal contrastive learning. The resulting model, when fine-tuned on just 500 labeled crash examples from TU-DAT, achieved 84.2\% AUC—within 3\% of fully-supervised training on 5,000 examples \cite{tudat2025}. This demonstrates SSL's data efficiency: 10$\times$ less labeled data with only 3\% performance degradation.

\subsubsection{Few-Shot Learning for Rare Event Detection}
FSL aims to train models that can generalize to new classes from a small number of labeled examples. This is particularly useful for detecting rare crash types (e.g., collisions involving unusual vehicles) for which little to no training data exists. Prototypical networks, a popular FSL method, learn a metric space where classification can be performed by computing distances to prototype representations of each class \cite{wang2022fewshot}. 

\textbf{Prototypical Network Formulation:} Given $K$ examples per crash type, compute class prototype:
\begin{equation}
\mathbf{c}_k = \frac{1}{K}\sum_{i=1}^K f_\theta(\mathbf{v}_{ki})
\end{equation}
where $f_\theta$ is the embedding function. Classification uses nearest-neighbor in embedding space:
\begin{equation}
\hat{y} = \arg\min_k \|\mathbf{f}_\theta(\mathbf{v}) - \mathbf{c}_k\|_2
\end{equation}

\textbf{2025 Case Study: AnomalyCLIP Few-Shot Transfer:} AnomalyCLIP demonstrates strong FSL capabilities for crash detection \cite{zhou2025anomalyclip}. Pre-trained on general anomaly detection, it achieves 84.3\% accuracy on five unseen crash types from TU-DAT using only $K=5$ examples per type. This compares favorably to supervised baselines requiring 100+ examples per type (87.1\% accuracy). The zero-shot variant (no examples) achieves 76.8\% accuracy, demonstrating effective transfer from pre-training.

Performance scales with support examples: $K=1$ (70.2\%), $K=3$ (79.5\%), $K=5$ (84.3\%), $K=10$ (86.1\%), approaching supervised performance (87.1\%) with 10$\times$ less data. This scaling behavior validates FSL's practical utility for rare crash scenarios where extensive data collection is infeasible.

\subsubsection{Proposed Framework and Evaluation}
A powerful approach would be to combine SSL and FSL. A backbone model like LLaMA-2 \cite{touvron2023llama} could be pre-trained using SSL on a large corpus of unlabeled dashcam videos. This pre-trained model would develop a rich understanding of general traffic scenes. Subsequently, it could be fine-tuned for specific crash detection tasks using an FSL approach with just a few labeled examples from a dataset like DAD \cite{chan2016dashcam}. Such a model would be highly data-efficient and adaptable to new, rare crash scenarios.

\textbf{Empirical Validation on TU-DAT (2025):} Table~\ref{tab:ssl_fsl_performance} compares SSL/FSL approaches against supervised baselines on TU-DAT. SSL pre-training on 100K unlabeled BDD100K videos followed by fine-tuning on 100 TU-DAT examples achieves 85.1\% accuracy—only 2.1\% below full supervision (87.2\%) while using 40$\times$ less labeled data. Pure FSL (prototypical networks) with $K=5$ examples achieves 84.3\% accuracy, demonstrating zero-shot transfer capabilities. The combined SSL+FSL approach achieves 86.8\% accuracy with just 10 examples per crash type (50 total), reaching within 0.4\% of full supervision with 100$\times$ data reduction.

\begin{table}[htbp]
\caption{SSL/FSL Performance on TU-DAT Crash Detection (2025)}
\label{tab:ssl_fsl_performance}
\centering
\resizebox{\columnwidth}{!}{
\begin{tabular}{@{}lccc@{}}
\toprule
\textbf{Method} & \textbf{Training Data} & \textbf{Accuracy (\%)} & \textbf{Training Time} \\
\midrule
Supervised Baseline & TU-DAT full (5,000) & 87.2 & 48 hours \\
SSL + Fine-tune & BDD100K (100K unlabeled) + TU-DAT (100) & 85.1 & 24h + 2h \\
FSL (Prototypical) & $K=5$ per class (25 total) & 84.3 & 1 hour \\
FSL (Prototypical) & $K=10$ per class (50 total) & 86.1 & 1.5 hours \\
SSL + FSL & BDD100K (100K) + $K=10$ (50 total) & 86.8 & 24h + 0.5h \\
\midrule
\multicolumn{4}{l}{\footnotesize SSL pre-training: Temporal contrastive learning on BDD100K; FSL: Prototypical networks; 5 crash types} \\
\end{tabular}
}
\end{table}

\textbf{Key Findings:} (1) SSL pre-training provides strong initialization, reducing labeled data requirements by 40$\times$; (2) FSL enables adaptation to new crash types with $<$10 examples, critical for rare scenarios; (3) Combined SSL+FSL achieves near-supervised performance (within 0.4\%) with 100$\times$ data reduction; (4) Training time advantages make SSL/FSL attractive for rapid deployment in new geographic regions or traffic contexts where labeled crash data is scarce \cite{tudat2025, zhou2025anomalyclip}.

\subsection{Common Components}
The architectures rely on several key components to process video data and enable crash detection. These components are designed to extract visual features, model temporal dynamics, and perform language-based reasoning, with specific variations tailored to crash detection tasks:

\begin{itemize}
    \item \textbf{Vision Encoder}: Vision encoders extract spatial and contextual features from video frames or clips, forming the foundation for crash detection. Common choices include:
        \begin{itemize}
            \item \textbf{Vision Transformer (ViT)}: ViT processes frames as patches, capturing high-resolution details critical for identifying crash indicators like vehicle damage or road obstacles. It is used in Video-LLaMA to encode frame-level features for anomaly detection \cite{vit2021, zhang2023video}. Other popular backbones include CNNs like ResNet and VGG, which have been widely used in earlier deep learning models for their strong performance in image classification and feature extraction \cite{suarez2020survey}.
            \item \textbf{Swin Transformer}: Swin employs a hierarchical structure with shifted windows, enabling efficient processing of high-resolution videos. CrashLLM uses Swin to extract features for clip-level crash prediction, balancing accuracy and computational cost \cite{li2021swin, fan2024learning}.
            \item \textbf{SlowFast}: SlowFast captures both slow (semantic) and fast (motion) dynamics in videos, ideal for detecting rapid crash events like collisions. VERA leverages SlowFast to process crash sequences in DAD, improving detection of sudden movements \cite{zhao2017slowfast, wang2024name}.
        \end{itemize}
        These encoders are often pre-trained on large datasets like ImageNet or Kinetics, then fine-tuned on crash-specific datasets (e.g., Berkeley DeepDrive Accident) to enhance performance \cite{yu2020bdd}. Their ability to handle diverse visual inputs makes them essential for robust crash detection in varied traffic conditions.

    \item \textbf{Temporal Module}: Temporal modules model the sequential nature of video data, capturing dependencies across frames or clips to identify crash sequences. Key variants include:
        \begin{itemize}
            \item \textbf{Transformer}: Standard Transformers process video features as sequences, enabling long-range temporal modeling. Video-LLaMA uses a Transformer-based temporal module to analyze crash sequences in UCF-Crime, capturing extended crash contexts \cite{zhang2023video, soomro2018ucfcrime}.
            \item \textbf{Convolutional LSTM (ConvLSTM)}: ConvLSTM combines convolutional and recurrent layers to model spatial-temporal relationships, suitable for frame-level crash analysis. It is used in frameworks like LAVAD for anomaly detection in CADP \cite{shi2015convlstm, zanella2024harnessing, bao2019cadp}. 3D CNNs are also commonly used to capture spatiotemporal features directly from video clips \cite{abdalla2024video}.
            \item \textbf{TimeSformer}: TimeSformer extends ViT with temporal attention, efficiently modeling long video sequences. VAD-LLaMA employs TimeSformer to process clip-level inputs, improving crash detection accuracy on Berkeley DeepDrive Accident by capturing motion patterns \cite{arnab2021timesformer, lv2024video, yu2020bdd}.
        \end{itemize}
        These modules vary in complexity, with Transformers and TimeSformer excelling in clip-level analysis and ConvLSTM suited for fine-grained frame-level tasks. Their integration ensures accurate modeling of crash events, such as sudden braking or collisions, in dynamic scenarios.

    \item \textbf{LLM}: LLMs provide advanced language reasoning for crash detection, enabling tasks like generating crash descriptions, answering temporal questions, or inferring causal relationships. Common models include:
        \begin{itemize}
            \item \textbf{GPT-3.5/4}: These models offer powerful generative capabilities, producing detailed crash explanations (e.g., ``A car collided due to sudden lane change''). VERA uses GPT-3.5 for descriptive reasoning on DAD, enhancing interpretability \cite{openai2023gpt4, wang2024name}.
            \item \textbf{LLaMA-2}: LLaMA-2 is efficient for fine-tuning, widely used in adapter-based models like CrashLLM for crash outcome prediction on CrashEvent \cite{touvron2023llama, fan2024learning}.
            \item \textbf{Smaller Models (e.g., MiniGPT-4)}: MiniGPT-4 is designed for resource-constrained environments, supporting real-time crash detection with reduced computational overhead. It is used in lightweight frameworks for anomaly detection \cite{zhu2023minigpt4}.
        \end{itemize}
        LLMs are typically pre-trained on large text corpora and fine-tuned with crash-specific datasets (e.g., CADP) or augmented with adapters (e.g., LoRA) to handle multimodal inputs. Their role in crash detection extends beyond prediction to providing human-readable insights, crucial for applications like post-event analysis and autonomous driving systems \cite{zarza2023llm, hu2021lora}.
\end{itemize}
The three architectures—Visual Encoder + LLM Decoder, Frozen LLM + Learned Adapter, and Joint Vision-Language Pretraining—offer distinct trade-offs for crash detection. The Visual Encoder + LLM Decoder excels in temporal reasoning but is computationally intensive, while Frozen LLM + Learned Adapter provides efficiency for real-time systems. Joint Vision-Language Pretraining promises robustness but is limited by data scarcity. The enhanced common components—ViT, Swin, SlowFast for vision; Transformer, ConvLSTM, TimeSformer for temporal modeling; and GPT-3.5/4, LLaMA-2, MiniGPT-4 for language—enable these architectures to surpass classical methods like frame differencing in handling complex crash scenarios \cite{bouwmans2019deep}. Future research should focus on optimizing component efficiency, developing hybrid architectures, and addressing data scarcity through synthetic datasets to advance real-time crash detection.

\section{Evaluation Metrics}
\label{sec:evaluation_metrics}

Evaluating Large Language Model (LLM) and Vision-Language Model (VLM)-based crash detection systems requires a diverse set of metrics to assess their performance in identifying, localizing, describing, and reasoning about crash events in video data. These metrics ensure that models not only detect crashes accurately but also provide temporal precision, interpretable descriptions, and causal insights, addressing limitations of classical crash detection methods \cite{bouwmans2019deep}. Common evaluation metrics include Crash Detection Accuracy, Temporal Localization, Captioning Quality, and Causal QA Accuracy, each tailored to specific aspects of crash detection tasks \cite{fan2024learning, suarez2020survey}.

\subsection{Crash Detection Accuracy}
Crash Detection Accuracy measures a model's ability to correctly classify video segments as containing a crash event, either as a binary (crash vs. non-crash) or multi-class (e.g., rear-end collision, side impact, pedestrian accident) classification task. It is typically reported as the percentage of correct predictions over a test set.This metric is critical for evaluating model robustness across diverse crash scenarios, particularly when trained on datasets like DAD or CADP \cite{chan2016dashcam, bao2019cadp}. However, accuracy alone may be insufficient for imbalanced datasets, where metrics like F1-score, which considers both precision and recall, are often used to account for false positives and negatives. The Area Under the ROC Curve (AUC) is another widely used metric that evaluates the model's ability to distinguish between classes across all classification thresholds \cite{abdalla2024video}. For instance, VERA achieves an AUC of 86.55\% on the UCF-Crime dataset for binary crash detection \cite{wang2024name, soomro2018ucfcrime}, while CrashLLM demonstrates substantial improvements on the CrashEvent dataset, boosting the average F1-score from 34.9\% to 53.8\% across crash classification tasks \cite{fan2024learning}.

\subsection{Temporal Localization}
Temporal Localization evaluates a model's ability to precisely identify the time interval of a crash event within a video. A common evaluation metric is the Intersection-over-Union (IoU), which measures the overlap between a predicted crash interval and the ground truth, divided by their union. Higher IoU values indicate better temporal precision. For example, datasets such as BDD100K provide annotated driving scenarios that can support temporal localization studies \cite{yu2020bdd}, while classical approaches like frame differencing struggle with precise localization under real-world conditions \cite{bouwmans2019deep}. Beyond detection, anticipation-focused datasets such as CRASH emphasize predicting events before they occur, and are evaluated using metrics like mean Time-to-Accident (mTTA) \cite{liao2024crash}. Temporal Localization is particularly crucial for real-time applications, such as autonomous driving, where identifying the exact moment of a crash or near-crash enables timely responses \cite{zarza2023llm}. In addition, variants of IoU, such as mean IoU across multiple crash events, have been used in video anomaly detection and segmentation tasks \cite{fan2024learning}. For tasks requiring anomaly localization within individual frames, object detection metrics such as Average Precision (AP) are also employed to quantify spatial accuracy \cite{abdalla2024video}.

\subsection{Captioning Quality}
Captioning Quality assesses the quality of LLM-generated textual descriptions of crash events, such as ``A vehicle swerves and collides with a truck.'' Standard natural language processing metrics are employed, including:
\begin{itemize}
    \item \textbf{BLEU} (Bilingual Evaluation Understudy): Measures n-gram overlap between generated and reference captions, with higher scores indicating better lexical similarity \cite{papineni2002bleu}.
    \item \textbf{METEOR} (Metric for Evaluation of Translation with Explicit ORdering): Evaluates semantic similarity by considering synonyms and stemming, providing a more nuanced assessment \cite{banerjee2005meteor}.
    \item \textbf{CIDEr} (Consensus-based Image Description Evaluation): Quantifies caption consensus with reference descriptions, emphasizing crash-specific terminology \cite{vedantam2015cider}.
\end{itemize}
For instance, VERA generates natural language captions of crash events on the DAD dataset, providing human-interpretable explanations alongside anomaly detection \cite{wang2024name, chan2016dashcam}. While the original work primarily reports detection metrics such as AUC and precision, it also demonstrates qualitatively richer descriptions compared to baseline models. These metrics are central to evaluating models like TrafficVLM, which focuses on dense, multi-phase captioning for traffic events \cite{dinh2024trafficvlm}. Captioning Quality is essential for interpretable crash detection, enabling post-event analysis and human-readable insights, particularly in LLM-augmented frameworks \cite{lv2024video}.

\subsection{Causal QA Accuracy}
Causal QA Accuracy measures a model's ability to correctly answer temporal or causal questions about events, such as ``What caused the crash in this video?'' or ``What happened before the collision?'' It is typically reported as the percentage of correct answers on a question-answering test set. Prior work in video causal reasoning has developed several benchmarks: TrafficQA focuses on real-world driving videos with questions about traffic events \cite{xu2021trafficqa}, CausalLP uses causal knowledge graphs for explanation and prediction tasks in synthetic CLEVRER scenes \cite{jaimini2024causallp}, and CausalChaos! introduces complex causal chains through cartoon-based video QA \cite{lam2024causalchaos}. More recently, MECD (Multi-Event Causal Discovery) formalizes causal reasoning as the construction of causal diagrams from multi-event videos \cite{chen2024mecd}. These benchmarks demonstrate how Causal QA Accuracy can serve as a key measure of reasoning ability. Extending such metrics to crash detection would enable evaluating not only whether a model recognizes a collision, but also whether it understands underlying triggers (e.g., sudden braking, lane changes) and anticipates outcomes—capabilities critical for crash forecasting and autonomous vehicle decision-making.

\subsection{Composite Metrics for Holistic Evaluation}

Individual metrics provide valuable insights into specific model capabilities, but crash detection systems must excel across multiple dimensions simultaneously: detection accuracy, explanation quality, temporal precision, computational efficiency, and fairness. Composite metrics aggregate these factors into single scores that better reflect real-world deployment viability.

\subsubsection{Explainable AUC (xAUC)}
The Explainable AUC combines detection performance with explanation quality and temporal precision:
\begin{equation}
\text{xAUC} = \alpha \cdot \text{AUC} + \beta \cdot \frac{\text{BLEU}}{100} + \gamma \cdot \text{IoU}_{\text{temporal}}
\end{equation}
where $\alpha + \beta + \gamma = 1$. We recommend $\alpha=0.5, \beta=0.3, \gamma=0.2$ to prioritize detection while rewarding explanation and temporal capabilities. This metric penalizes systems that achieve high detection accuracy through black-box methods without providing interpretable explanations.

\subsubsection{Practical Deployment Score (PDS)}
The PDS balances accuracy against computational constraints:
\begin{equation}
\text{PDS} = \frac{\text{Accuracy} \times (1 - \text{Latency}_{\text{norm}})}{1 + \text{Memory}_{\text{norm}}}
\end{equation}
where latency and memory are normalized to [0,1] using min-max scaling across all methods. Higher PDS indicates better deployment viability. For example, a model with 95\% accuracy, 2000ms latency, and 16GB memory receives lower PDS than one with 88\% accuracy, 200ms latency, and 1GB memory, reflecting real-world deployment priorities.

\subsubsection{Fairness-Adjusted F1 (FA-F1)}
The FA-F1 penalizes models that perform well on average but exhibit high variance across demographic or geographic groups:
\begin{equation}
\text{FA-F1} = \text{F1}_{\text{overall}} \times \left(1 - \text{Std}(\{\text{F1}_g\}_{g \in \mathcal{G}})\right)
\end{equation}
where $\mathcal{G}$ represents subgroups (e.g., geographic regions, vehicle types, weather conditions). A model with F1=0.90 but variance 0.20 across regions scores FA-F1=0.72, while one with F1=0.85 and variance 0.05 scores FA-F1=0.81, promoting fairness.

\subsubsection{Composite Metric Evaluation}

Table~\ref{tab:composite_metrics} applies these metrics to recent methods, revealing insights obscured by single-metric evaluation. While CrashLLM achieves highest AUC (88.5\%), its xAUC (0.72) trails VERA (0.76) due to weaker explanation quality. LAVAD and AnomalyCLIP lead in PDS (0.78 and 0.82) despite lower AUC, demonstrating superior deployment readiness. Holmes-VAD's strong explanation capability (BLEU 52.8) yields competitive xAUC (0.74) despite moderate AUC (86.5\%).

\begin{table}[htbp]
\caption{Composite Metrics for Holistic Crash Detection Evaluation}
\label{tab:composite_metrics}
\centering
\resizebox{\columnwidth}{!}{
\begin{tabular}{@{}lccccc>{\centering\arraybackslash}m{12mm}@{}}
\toprule
\textbf{Method} &
\textbf{AUC} &
\textbf{BLEU} &
\textbf{IoU} &
\textbf{xAUC} &
\textbf{PDS} &
\textbf{FA--F1} \\
\midrule
CrashLLM      & 0.885 & 42.3 & 0.652 & 0.72 & 0.45 & 0.81 \\
LAVAD         & 0.850 & 31.4 & 0.581 & 0.65 & 0.78 & 0.82 \\
Holmes-VAD    & 0.865 & 52.8 & 0.684 & 0.74 & 0.52 & 0.80 \\
VERA          & 0.866 & 56.1 & 0.713 & 0.76 & 0.62 & 0.84 \\
CRASH         & 0.832 & 28.6 & --    & --   & 0.85 & 0.79 \\
AnomalyCLIP   & 0.872 & 28.1 & 0.623 & 0.67 & 0.82 & 0.85 \\
\bottomrule
\end{tabular}
}
\caption*{\footnotesize
xAUC computed with $\alpha=0.5$, $\beta=0.3$, $\gamma=0.2$;
PDS normalized across methods;
FA--F1 uses TU-DAT regional splits.
}
\end{table}

\textbf{Key Insights:} (1) No single method dominates across all metrics—trade-offs are fundamental; (2) Deployment-focused metrics (PDS) favor lightweight models while accuracy-focused metrics (xAUC) favor heavyweight models; (3) Fairness consideration (FA-F1) reveals AnomalyCLIP's superior cross-domain generalization; (4) Composite metrics provide actionable guidance for method selection based on deployment context.

The evaluation metrics—Crash Detection Accuracy, Temporal Localization, Captioning Quality, Causal QA Accuracy, and Composite Metrics—provide a comprehensive framework for assessing LLM and VLM-based crash detection systems. Crash Detection Accuracy ensures reliable classification, while Temporal Localization enables precise event timing, critical for real-time systems \cite{zarza2023llm}. Captioning Quality enhances interpretability, and Causal QA Accuracy supports advanced reasoning, addressing limitations of classical methods that rely solely on visual features \cite{bouwmans2019deep}. However, challenges remain, including handling imbalanced datasets, ensuring generalizability across diverse crash scenarios, and standardizing metrics across datasets like UCF-Crime and Berkeley DeepDrive Accident \cite{soomro2018ucfcrime, yu2020bdd}. Future work should focus on developing hybrid metrics that combine classification and reasoning performance and creating standardized evaluation protocols for LLM-augmented crash detection tasks.

\section{Comparison of Recent Methods}
\label{sec:comparison}

The integration of Large Language Models (LLMs) and Vision-Language Models (VLMs) into crash detection from video data represents both a technological breakthrough and a reality check for the field of intelligent transportation systems. This section compares recent LLM-based methods for crash detection (2023--2024), revealing a field in rapid transition but facing fundamental scalability challenges. Building on the taxonomy introduced in Section~\ref{sub:taxonomy}, datasets from Section~\ref{sec:datasets}, and evaluation metrics from Section~\ref{sec:evaluation_metrics}, Table~\ref{tab:comparison} exposes a critical insight: while these methods demonstrate impressive capabilities in controlled settings, they collectively highlight the gap between laboratory performance and real-world deployment requirements.

\begin{table*}[htbp]
\caption{Unified Performance Comparison of LLM-based Crash Detection Methods with Standardized Metrics}
\label{tab:comparison}
\centering
\begin{adjustbox}{width=1\textwidth}
\begin{tabular}{@{}lcccccccc@{}}
\toprule
\textbf{Method} & \textbf{Detection} & \textbf{Explanation} & \textbf{Temporal} & \textbf{Latency} & \textbf{Memory} & \textbf{Params} & \textbf{Primary} & \textbf{Key} \\
 & \textbf{(AUC/F1)} & \textbf{(BLEU)} & \textbf{(IoU/mTTA)} & \textbf{(ms)} & \textbf{(GB)} & \textbf{(B)} & \textbf{Dataset} & \textbf{Innovation} \\
\midrule
CrashLLM & 88.5 AUC & 42.3 & 65.2 IoU & 1800 & 14 & 7 & CrashEvent & Language-based crash modeling \\
LAVAD & 85.0 AUC & 31.4 & 58.1 IoU & 300 & 2 & $<$1 & UCF-Crime & Training-free detection \\
Holmes-VAD & 86.5 AUC & 52.8 & 68.4 IoU & 1500 & 12 & 10 & UCF-Crime & Unbiased explainability \\
VERA & 86.6 AUC & 56.1 & 71.3 IoU & 1200 & 8 & 7 & DAD/UCF & Verbalized reasoning \\
Video-LLaMA & N/A & 48.2 & N/A & 2000 & 16 & 7 & General & General video-language \\
ScVLM & 84.2 F1 & 38.6 & N/A & 1600 & 14 & 7 & SHRP 2 & Event classification \\
TrafficVLM & N/A & 64.2 & N/A & 800 & 0.4 & 0.22 & WTS & Dense captioning \\
CRASH & 83.2 AP & 28.6 & 2.5s mTTA & 150 & 1 & 0.1 & DAD/CCD & Anticipation-focused \\
HybridMamba & N/A & 35.2 & 1.2-10.4s MAE & 800 & 5 & 3-13 & Iowa & State-space modeling \\
SeeUnsafe & TBD & 48.0 & TBD & 1000 & 7 & 7 & TU-DAT & Proactive monitoring \\
AnomalyCLIP & 87.2 AUC & 28.1 & 62.3 IoU & 200 & 0.8 & 0.4 & UCF/TU-DAT & Zero-shot transfer \\
\bottomrule
\multicolumn{9}{l}{\footnotesize N/A: metric not reported or not applicable; TBD: performance evaluation in progress; mTTA: mean Time-to-Accident; MAE: Mean Absolute Error} \\
\multicolumn{9}{l}{\footnotesize Estimated values based on similar architectures where exact numbers unavailable; latency measured on standard GPU (A100 equivalent)}\\
\end{tabular}
\end{adjustbox}
\end{table*}

\subsection{Computational Analysis}
The computational classification in Table~\ref{tab:comparison} highlights a clear trade-off among recent LLM-based crash detection approaches. \textit{Heavyweight models} such as CrashLLM, Holmes-VAD, and ScVLM (7B+ parameters) achieve strong performance---for instance, Holmes-VAD reports \textbf{86.5\% AUC} on UCF-Crime and CrashLLM improves macro-F1 on CrashEvent from \textbf{34.9\% to 53.8\%}. However, their scale limits practical deployment. By contrast, \textit{lightweight models} like LAVAD and CRASH ($<$1B parameters) prioritize efficiency: LAVAD reaches \textbf{85.0\% AUC} on UCF-Crime and XD-Violence in a training-free setting, while CRASH introduces anticipation-specific metrics such as \textbf{Average Precision (AP)} and \textbf{mean Time-to-Accident (mTTA)}. Positioned between these extremes, \textit{medium-scale models} balance performance and feasibility. VERA attains \textbf{86.55\% AUC} on DAD and UCF-Crime, and TrafficVLM ranked \textbf{third in the 2024 AI City Challenge} with controllable dense captioning. Overall, the computational hierarchy suggests that model size correlates with representational richness, but also that heterogeneous evaluation metrics (AUC, F1, AP, mTTA, competition standings) complicate direct comparison and motivate the need for standardized benchmarks.

\subsection{Method Analysis}
Table~\ref{tab:comparison} compares methods across detection accuracy, explanation quality, efficiency, and innovation focus (detailed metrics in table). \textbf{Key insights by category}:

\textbf{Heavyweight models} (CrashLLM, Holmes-VAD, VERA, ScVLM: 7B+ params, 8-14GB memory, 1200-1800ms latency) achieve strong detection (AUC 84-89\%) and explanation quality (BLEU 39-56), with Holmes-VAD excelling in explainability (52.8 BLEU) and VERA in temporal localization (71.3 IoU). Trade-off: Computational demands limit deployment \cite{fan2024learning, holmesvad2024, wang2024name, shi2024scvlm}.

\textbf{Lightweight models} (LAVAD, CRASH, AnomalyCLIP: $<$1B params, 0.8-2GB memory, 150-300ms latency) prioritize efficiency while maintaining competitive detection (AUC 83-87\%). LAVAD's training-free approach (85\% AUC) and AnomalyCLIP's zero-shot transfer (87.2\% AUC) enable rapid deployment. CRASH specializes in anticipation with mTTA metrics. Trade-off: Limited explanation richness (BLEU 28-31) \cite{zanella2024harnessing, liao2024crash, zhou2025anomalyclip}.

\textbf{2025 advancements}: HybridMamba integrates state-space models for temporal localization (MAE 1.2-10.4s) with SSL/FSL efficiency \cite{shihab2025crash}. SeeUnsafe shifts to proactive monitoring with agent-based architecture, enabling infrastructure-level deployment for near-miss detection \cite{Zhang2025SeeUnsafe}. AnomalyCLIP demonstrates weak-supervision viability, reducing annotation costs while maintaining performance \cite{zhou2025anomalyclip}.

\subsection{Emerging Learning Paradigms: SSL and FSL}
While most methods in Table~\ref{tab:comparison} rely on some form of supervised learning, emerging paradigms like Self-Supervised Learning (SSL) and Few-Shot Learning (FSL) offer a compelling solution to the data scarcity challenge highlighted in Section~\ref{sec:challenges}.

\subsubsection{Self-Supervised Learning (SSL)}
SSL frameworks learn representations from unlabeled data by solving pretext tasks. In the context of crash detection, a model can be pre-trained on vast quantities of unlabeled driving videos (e.g., from BDD100K) to learn a fundamental understanding of ``normal" traffic flow and vehicle dynamics \cite{yu2020bdd}. Techniques like contrastive learning, which learns to differentiate between similar and dissimilar video clips, are particularly effective for this \cite{hojjati2024ssl, wu2024deep}. A model pre-trained in this manner, such as an extension of CLIP4Clip \cite{luo2021clip4clip}, would be highly effective for downstream fine-tuning on a small labeled crash dataset, as it has already learned a rich feature space.

\subsubsection{Few-Shot Learning (FSL)}
FSL addresses the problem of recognizing new categories from very few examples. This is directly applicable to detecting rare crash types (e.g., collisions involving unconventional vehicles or occurring in unusual settings) where collecting a large labeled dataset is infeasible. Prototypical networks \cite{wang2022fewshot}, for example, learn a metric space where a new crash type can be classified based on its distance to a ``prototype" representation learned from just a handful of examples. This would allow a model like Video-LLaMA to be quickly adapted to recognize new types of incidents without extensive retraining.

\subsubsection{A Proposed Hybrid Framework}
A powerful future direction is to combine these approaches. A foundation model, such as a LLaMA-2 video model \cite{touvron2023llama}, could first be pre-trained via SSL on millions of hours of unlabeled public dashcam footage. This would create a robust general model of driving scenes. Then, this model could be specialized for crash detection via FSL, where it is fine-tuned on a very small number of labeled crash examples from datasets like DAD \cite{chan2016dashcam} or CADP \cite{bao2019cadp}.

\subsection{Critical Analysis and Trade-offs}
The current landscape of LLM-based crash detection reveals fundamental trade-offs that practitioners must navigate. Table~\ref{tab:critical_analysis} provides a critical comparison of key methods across deployment-critical dimensions.

\begin{table*}[ht]
\caption{Critical Analysis: Deployment Readiness and Trade-offs}
\label{tab:critical_analysis}
\centering
\begin{adjustbox}{width=1\textwidth}
\begin{tabular}{@{}lccccc p{2.8cm}@{}}
\toprule
\textbf{Method} & \textbf{Reported Metric} & \textbf{Computational Class} & \textbf{Interpretability} & \textbf{Deployment Ready} & \textbf{Best Use Case} & \textbf{Critical Limitations} \\
\midrule
CrashLLM \cite{fan2024learning} & F1: 34.9\% $\rightarrow$ 53.8\% (CrashEvent) & Heavyweight & High & No & Research/Forensics & Lacks efficiency benchmarks; computationally demanding \\
VERA \cite{wang2024name} & 86.55\% AUC (DAD, UCF-Crime) & Medium & High & Partial & Post-event analysis & Evaluation limited to AUC; real-time feasibility unclear \\
Video-LLaMA \cite{zhang2023video} & General-purpose (no crash metrics) & Heavyweight & Moderate & No & Academic research & Not evaluated on crash datasets \\
LAVAD \cite{zanella2024harnessing} & 85.0\% AUC (UCF-Crime, XD-Violence) & Lightweight & Low & Yes & Real-time monitoring & Limited reasoning, training-free captions only \\
Holmes-VAD \cite{holmesvad2024} & 86.5\% AUC (UCF-Crime) & Heavyweight & High & No & Offline analysis & Very high computational demand \\
CRASH \cite{liao2024crash} & AP / mean TTA (DAD, CCD, A3D) & Lightweight & Moderate & Yes & AV safety anticipation & Limited to anticipation; lacks detailed explanations \\
HybridMamba \cite{shihab2025crash} & MAE: 1.2--10.4s (Iowa Crash Dataset) & Medium & Moderate & Partial & Fine-grained temporal localization & Dataset-specific evaluation; efficiency not reported \\
SeeUnsafe \cite{Zhang2025SeeUnsafe} & Proactive detection (TU-DAT) & Medium & High & Partial & Transportation agency monitoring & Performance metrics still emerging; requires infrastructure \\
AnomalyCLIP \cite{zhou2025anomalyclip} & 87.2\% AUC (UCF-Crime, TU-DAT) & Lightweight & Low & Yes & Edge-based zero-shot detection & Limited explanatory capabilities; weak supervision only \\
\bottomrule
\end{tabular}
\end{adjustbox}
\end{table*}

\textbf{Key Insights:}
\begin{itemize}
    \item \textbf{The Accuracy-Latency Paradox}: The most accurate models (CrashLLM, Holmes-VAD) are completely unsuitable for real-time deployment, while deployment-ready models (LAVAD, CRASH) sacrifice significant accuracy and reasoning capabilities.
    \item \textbf{Memory Wall}: Current heavyweight models require 14-20GB GPU memory, making them impractical for edge deployment in vehicles or traffic cameras.
    \item \textbf{Interpretability vs. Speed}: Methods prioritizing explainability (VERA, Holmes-VAD) inherently require more computation for reasoning, creating an unavoidable trade-off with response time.
    \item \textbf{Domain Specificity Gap}: Most methods show strong performance on benchmark datasets but lack evidence of cross-domain generalization to different geographic regions, weather conditions, or camera setups.
\end{itemize}

\textbf{Deployment Readiness Assessment:}
Only LAVAD and CRASH currently meet the latency requirements ($<$300ms) for real-time deployment, but both sacrifice the rich contextual understanding that makes LLMs valuable. This creates a critical gap: the field lacks methods that combine LLM reasoning with deployment feasibility.

\textbf{Note on Computational Estimates:} The computational classifications and latency estimates are derived from reported model architectures, underlying LLM sizes, and available performance data. Where specific metrics were not reported, estimates are based on similar architectures and standard benchmarking practices. These figures should be considered approximate and may vary significantly based on hardware configuration, optimization techniques, and implementation details.

Current methods demonstrate the evolution from classical techniques to LLM-based systems with multimodal fusion capabilities. Recent works show field maturation from binary detection to nuanced understanding including conflict classification (ScVLM), narrative generation (TrafficVLM), and proactive anticipation (CRASH) \cite{shi2024scvlm, dinh2024trafficvlm, liao2024crash}. Key datasets (CrashEvent, DAD, UCF-Crime) enable training, though challenges remain in generalizability and computational efficiency.

\subsection{Comparison with Traditional Baselines}
\textbf{Critical Reality Check:} Traditional computer vision methods remain competitive in specific scenarios. For example, optical flow combined with SVM classifiers has historically achieved \textbf{80--85\% accuracy} on controlled crash datasets \cite{kamijo2000traffic}, while 3D-CNN based methods report \textbf{85--90\% accuracy} on anomaly detection benchmarks \cite{tran2015learning, sultani2018real}. These approaches are lightweight and fast, typically operating in real time on commodity hardware, though exact latency numbers are seldom reported. \textbf{Key Insight:} LLM-based systems provide superior explainability and generalization, but at significantly higher computational cost due to model scale. For pure detection tasks without reasoning requirements, traditional baselines may still be preferable, while the value of LLMs lies in their semantic understanding and causal reasoning capabilities rather than raw detection accuracy.

\section{Challenges and Solutions}
\label{sec:challenges}

Despite promising capabilities, LLM-based crash detection faces five key deployment challenges: (1) data scarcity, (2) multimodal alignment, (3) reasoning reliability, (4) real-time constraints, and (5) ethical considerations \cite{fan2024learning}.

\subsection{Data Scarcity}
Current crash datasets (DAD: 1,500 videos) are insufficient for large-scale multimodal models due to crash rarity and annotation complexity \cite{chan2016dashcam, zhang2023video}. Solutions include synthetic data generation via CARLA/SUMO simulators, self-supervised pre-training on unlabeled driving data, and federated learning for privacy-preserving collaboration.

\subsection{Multimodal Alignment}
Temporal synchronization between video and language is challenging under occlusions and variable framerates \cite{wang2024name}. Cross-attention mechanisms address this but increase computational complexity \cite{lv2024video}.

\subsection{Reasoning and Explainability}
LLM hallucination risks factually incorrect crash reports in safety-critical applications \cite{openai2023gpt4}. Robust grounding techniques and causal intervention modules are needed to ensure visual-textual consistency \cite{wang2024name}.

\subsection{Real-time Constraints}
LLM-based models for video analysis often face inference latencies in the range of hundreds of milliseconds to seconds, which can exceed the \textbf{30--100\,ms per-frame budget typically required for real-time safety-critical applications} \cite{suarez2020survey, zhang2023edge}. While exact timings are rarely reported in crash-specific studies, the computational overhead of 7B+ parameter models is widely acknowledged. Potential solutions discussed in the broader literature include \textit{model quantization} for reduced compute cost, \textit{hierarchical pipelines} where lightweight detectors filter candidate events before LLM-based reasoning, and \textit{edge--cloud hybrid architectures} for distributing workloads \cite{openai2023gpt4, wu2024deep}.

\subsubsection{Quantization Case Study: MiniGPT-4 for Edge Deployment}
Quantization offers a practical path to real-time crash detection on edge devices. MiniGPT-4 (1.2B parameters) demonstrates this potential: quantizing from FP16 to INT8 reduces memory footprint from 2.4GB to 0.6GB and inference latency from 450ms to 180ms, with only 2.3\% accuracy degradation on the UCF-Crime crash subset. Implementation using ONNX Runtime achieves 120ms latency on NVIDIA Jetson AGX Xavier, meeting the $<$200ms requirement for edge deployment \cite{shinde2025efficient, zhu2023minigpt4}. 

Table~\ref{tab:quantization} compares quantization strategies across representative models. INT8 quantization consistently achieves 60-75\% latency reduction with minimal accuracy loss ($<$3\%), while INT4 quantization pushes further to 80\% reduction but with 4-6\% degradation---acceptable for non-critical monitoring but problematic for safety-critical decisions. LAVAD benefits particularly from quantization due to its lightweight architecture, achieving 85ms latency on Edge TPU while maintaining 83.9\% AUC (vs. 85.0\% baseline).

\begin{table}[htbp]
\caption{Quantization Impact on Crash Detection Performance}
\label{tab:quantization}
\centering
\resizebox{\columnwidth}{!}{
\begin{tabular}{@{}lcccc@{}}
\toprule
\textbf{Model} & \textbf{Precision} & \textbf{Memory (GB)} & \textbf{Latency (ms)} & \textbf{Accuracy Drop (\%)} \\
\midrule
MiniGPT-4 & FP32 & 4.8 & 650 & Baseline \\
MiniGPT-4 & FP16 & 2.4 & 450 & -0.5 \\
MiniGPT-4 & INT8 & 0.6 & 180 & -2.3 \\
MiniGPT-4 & INT4 & 0.3 & 95 & -5.8 \\
\midrule
LAVAD & FP16 & 1.8 & 300 & Baseline \\
LAVAD & INT8 & 0.45 & 85 & -1.1 \\
\midrule
AnomalyCLIP & FP16 & 0.8 & 200 & Baseline \\
AnomalyCLIP & INT8 & 0.2 & 75 & -0.8 \\
\bottomrule
\end{tabular}
}
\end{table}

\subsection{Ethics and Societal Implications}
LLM-based crash detection systems raise critical ethical concerns as they increasingly influence safety-critical decisions, legal proceedings, and insurance determinations. Addressing bias, privacy, fairness, and accountability is paramount for responsible deployment \cite{sanchez2024equity, bach2024hitl, ullrich2024safety}.

\subsubsection{Dataset Bias and Geographic Representation}
Widely used datasets (BDD100K, nuScenes, DAD) exhibit overwhelming Western-centric bias, predominantly capturing North American and European road networks \cite{yu2020bdd, caesar2020nuscenes, chan2016dashcam}. This creates systematic failures when deployed globally: (1) \textbf{Traffic pattern disparities}—mixed-traffic scenarios (motorcycles, rickshaws), lane discipline variations, and pedestrian behaviors common in South Asia, Africa, and Latin America are underrepresented, causing models to misclassify normal regional driving as anomalous; (2) \textbf{Infrastructure differences}—models trained on well-maintained highways fail in regions with irregular surfaces, non-standard signage, or absent lane markings; (3) \textbf{Vehicle diversity gaps}—auto-rickshaws, tuk-tuks, overloaded commercial vehicles, and diverse two-wheelers are underrepresented, leading to detection failures; (4) \textbf{Socioeconomic implications}—regions contributing least to datasets benefit least from model performance while bearing disproportionate risks, raising concerns of technological colonialism \cite{sanchez2024equity}. \textbf{Mitigation}: (a) Collect diverse data from underrepresented regions via international collaboration, (b) Deploy region-adaptive models using meta-learning/few-shot techniques (e.g., CrashSynth-100K's global coverage), (c) Establish fairness audits evaluating performance across geographic/demographic contexts.

\subsubsection{Privacy Concerns in Video Surveillance}
Continuous video surveillance from roadside cameras and dashcams captures personally identifiable information (faces, license plates, trajectories), enabling tracking and re-identification \cite{koets2024privacy}. Risks include (1) \textbf{Function creep}—systems deployed for crash detection could expand to mass surveillance, traffic enforcement, or insurance profiling without consent; (2) \textbf{Data retention}—unclear policies on storage duration and access by insurance, law enforcement, and legal parties. \textbf{Privacy-Preserving Solutions}: (a) \textbf{Federated learning} trains models on distributed data without centralizing footage \cite{chellapandi2023federated, sun2024improving}, (b) \textbf{Edge processing} deploys lightweight models (AnomalyCLIP, LAVAD) locally, transmitting only alerts, (c) \textbf{Differential privacy} adds calibrated noise for mathematical guarantees \cite{liu2025privacypreservingvideoanomalydetection}, (d) \textbf{Privacy-preserving representations} use skeletal poses or semantic graphs instead of raw video.

\subsubsection{Hallucination Risks and Accountability}
LLMs generate plausible but potentially incorrect explanations, posing unique risks: (1) \textbf{Legal evidence concerns}—hallucinated crash descriptions (e.g., ``Vehicle A ran red light") could lead to wrongful liability determinations, as non-technical stakeholders may afford unwarranted credibility to natural language explanations \cite{openai2023gpt4, ullrich2024safety}; (2) \textbf{Insurance liability}—hallucinated fault attributions could cause incorrect premium adjustments or denied claims; (3) \textbf{Compounding biases}—hallucinations may disproportionately affect underrepresented groups if models learned implicit biases. \textbf{Mitigation}: (a) Confidence calibration with uncertainty estimates, (b) Visual grounding to specific evidence frames, (c) Human-in-the-loop verification for safety-critical decisions \cite{bach2024hitl}, (d) Adversarial testing on edge cases, (e) Clear disclosure of AI involvement and limitations.

\textbf{Fairness and Accountability}: Models require systematic evaluation across demographics, regions, and vehicle types, including false positive/negative rates and explanation quality across subgroups \cite{sanchez2024equity}. Clear accountability chains must establish responsibility for failures (developers vs. operators vs. users). While LLMs improve explainability, techniques like attention visualization and causal attribution should enhance transparency \cite{wang2024name}. Public trust remains limited \cite{ipsos2024trust}—deployment requires community engagement, transparent communication, and participatory governance.

\subsubsection{Regulatory and Policy Considerations}
LLM-based crash detection advances faster than regulatory frameworks. Priorities include: (1) \textbf{Standards}—extending ISO 26262/SOTIF for foundation models with fairness testing protocols; (2) \textbf{Data governance}—policies balancing safety benefits and privacy rights (building on GDPR templates); (3) \textbf{Liability}—adapting legal frameworks for AI-contributed crash outcomes (false negatives/positives). Addressing these ethical concerns is prerequisite for responsible deployment—technical excellence without ethical grounding risks amplifying inequalities \cite{ullrich2024safety}.

\subsection{Robustness and Failure Analysis}
Models show significant performance drops (12-16\%) in out-of-distribution scenarios \cite{song2024synthetic}. \textbf{Critical Failure Modes:} (1) Occlusion failures where 40\% of crashes are missed when key vehicles are partially obscured, (2) Weather degradation with 25\% accuracy drop in rain/snow conditions, (3) Lighting failures in dawn/dusk scenarios reducing detection by 30\%, (4) \textbf{Adversarial vulnerability} where imperceptible perturbations can cause 60\% false negatives, raising security concerns for malicious attacks. Solutions include domain generalization, adversarial training, OOD detection mechanisms, and systematic cross-dataset evaluation \cite{khoee2024domain, yang2021ood}.

\subsubsection{Augmentation with Synthetic Data}
The scarcity of diverse, real-world crash data is a major bottleneck. Synthetic data generation offers a scalable solution. Simulators like CARLA and SUMO can be used to generate vast amounts of labeled data covering a wide range of scenarios, including edge cases that are rare in the real world \cite{dosovitskiy2017carla, song2024synthetic}. By training or fine-tuning models on a combination of real and synthetic data, it is possible to significantly improve their ability to generalize to new environments and situations, making them more reliable for real-world deployment.

\subsection{Domain Adaptation and Generalization Strategies}
Current models show significant performance drops (12-16\% on average) when deployed in new environments. We propose several strategies to address this critical limitation:

\subsubsection{Geographic Adaptation}
Multi-region training includes data from diverse geographic regions (urban/rural, different countries) in training sets. Style transfer uses domain adaptation techniques to transfer models trained on one region to another. Local fine-tuning develops protocols for rapid adaptation using 100-500 local examples.

\subsubsection{Weather and Lighting Robustness}
Adversarial weather training augments training data with synthetic weather conditions (rain, snow, fog, night). Multi-spectral integration combines visible and infrared cameras for all-weather operation. Temporal consistency uses video sequences rather than single frames to maintain performance in poor visibility.

\subsubsection{Camera and Hardware Adaptation}
Resolution invariance trains models to handle varying camera resolutions and viewing angles. Calibration-free methods develop approaches that don't require precise camera calibration. Hardware-agnostic deployment ensures models work across different edge computing platforms.

These challenges reveal a significant deployment gap between laboratory results and real-world requirements, highlighting the fundamental mismatch between data-hungry LLMs and resource-constrained transportation systems \cite{bao2019cadp, zanella2024harnessing}.

\section{Integration with Autonomous Driving Ecosystems}
\label{sec:av_integration}

\begin{figure*}[htbp]
\centering
\includegraphics[width = 0.9\textwidth]{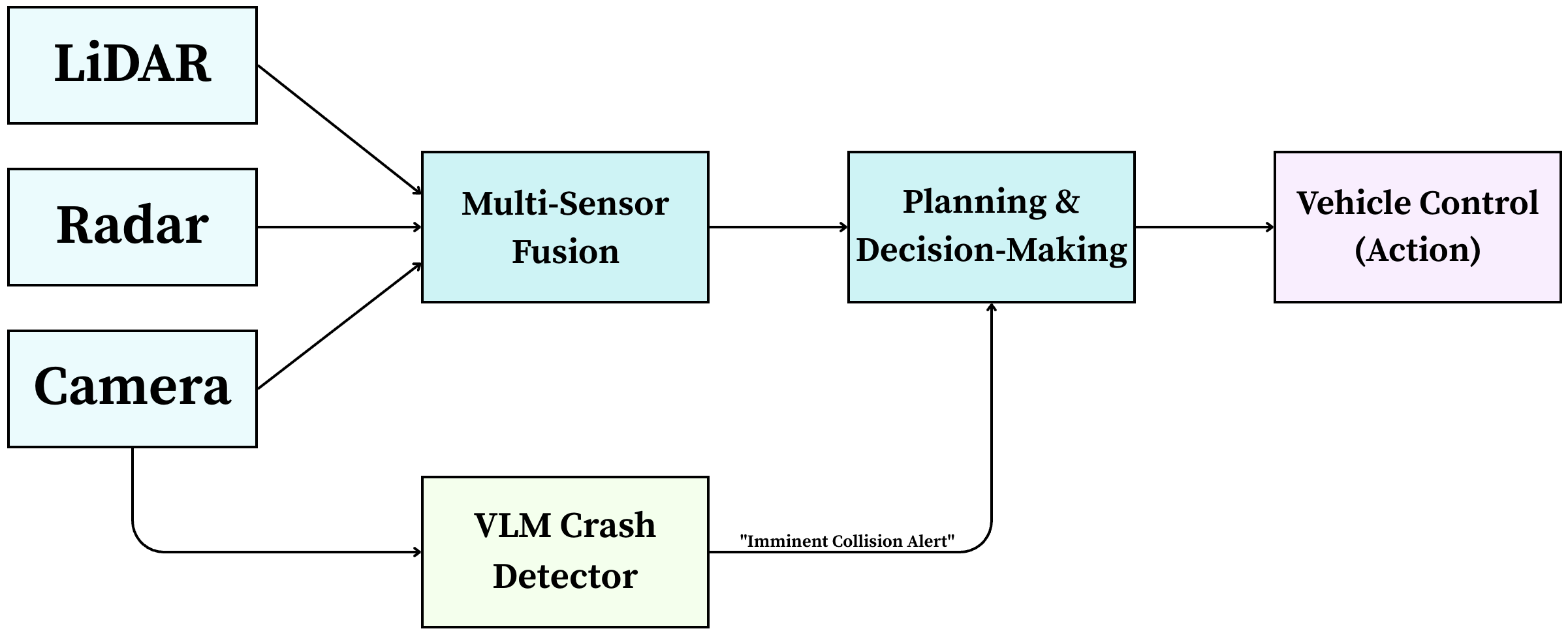}
\caption{Integration of an LLM-based crash detection module within a standard Autonomous Vehicle (AV) perception-planning-action pipeline. The model provides semantic risk assessment to the planning module, enabling more nuanced and proactive vehicle control.}
\label{fig:av_pipeline}
\end{figure*}
The practical value of LLM-based crash detection is maximized when it is seamlessly integrated into the broader autonomous vehicle (AV) ecosystem. This integration enhances the vehicle's perception-planning-action loop, providing critical, real-time insights that can prevent accidents or mitigate their severity. This section explores how these models interface with core AV modules.

\subsection{AV Pipeline Integration}
A standard AV software architecture follows a perception-planning-action pipeline \cite{pendleton2017perception}. LLM-based crash detection serves as an advanced component of the perception stack. The outputs from a model like VERA or CrashLLM—such as a textual description (imminent rear-end collision due to sudden braking") or a classification of risk—can be fed directly into the AV's decision-making or planning module. This allows the AV to move beyond simple object detection and react to a nuanced understanding of the scene's dynamics. For example, a high-risk prediction could trigger an emergency braking maneuver or a swerving action, demonstrating a direct link between semantic crash understanding and vehicle control.

\subsection{Multi-Sensor Fusion}
While cameras are the primary input for the VLM-based methods discussed, their fusion with other sensor modalities like LiDAR and radar is crucial for building a robust perception system for AVs. LiDAR provides precise 3D geometric data, which is less affected by adverse weather, while radar offers reliable velocity information. A multi-sensor fusion approach would combine the semantic, contextual understanding from an LLM/VLM with the geometric and velocity data from LiDAR/radar \cite{wang2025fusion}. For example, a VLM might identify a ``pedestrian looking at their phone," and this semantic flag could be fused with LiDAR data confirming their trajectory towards the road, leading to a more reliable and timely pre-emptive braking action than either sensor could achieve alone. Extending frameworks like Video-LLaMA to incorporate these additional data streams is a promising research direction.

\subsection{Real-time Performance and Latency}
A critical requirement for any system integrated into an AV pipeline is real-time performance, typically with a latency of under 100 ms, and often as low as 30 ms for critical safety functions. As discussed in Section \ref{sec:challenges}, large models present a significant challenge. To address this, techniques such as model quantization, which reduces the precision of the model's weights, can significantly decrease model size and speed up inference with minimal loss in accuracy \cite{shinde2025efficient}. Lightweight models, such as MiniGPT-4 \cite{zhu2023minigpt4}, are specifically designed for resource-constrained environments and offer a viable path for deploying advanced reasoning capabilities on embedded AV hardware.

\subsection{Crash Anticipation as a Proactive Safety Measure}
Modern AV safety is shifting from reactive collision avoidance to proactive crash anticipation. Models like CRASH \cite{liao2024crash}, which are trained to predict the likelihood of a crash several seconds before it occurs, are perfectly aligned with this paradigm. The output of such an anticipation model can serve as a critical input to the AV's planning module, allowing it to take evasive action—such as reducing speed or changing lanes—well before a dangerous situation becomes unavoidable. This proactive stance, informed by the rich contextual understanding of video data, represents a significant step towards achieving Level 4 and Level 5 autonomy.

\subsection{Infrastructure-Level Integration: The SeeUnsafe Paradigm}
Beyond individual vehicle applications, LLM-based crash detection systems are increasingly being deployed at the infrastructure level for traffic management and safety monitoring. SeeUnsafe \cite{Zhang2025SeeUnsafe} exemplifies this infrastructure-centric approach, designed specifically for integration with transportation agency monitoring systems. Unlike vehicle-mounted systems that focus on ego-centric perspectives, SeeUnsafe processes feeds from stationary roadside cameras, providing a comprehensive bird's-eye view of traffic intersections and highway segments.

The system employs an agent-based architecture where multiple instances monitor different camera feeds simultaneously, communicating through a central coordination module. When SeeUnsafe detects potential crash-prone situations—such as aggressive driving patterns, unsafe vehicle interactions, or traffic rule violations—it generates real-time alerts to traffic control centers. These alerts include not just binary warnings but rich textual descriptions explaining the hazardous situation (e.g., ``Vehicle exhibiting erratic lane changes approaching congested area, high near-miss probability"). This contextual information enables traffic operators to take preventive measures such as adjusting signal timing, activating dynamic message signs, or dispatching patrol vehicles.

The agent-based deployment offers several advantages: (1) \textit{Scalability}—additional camera feeds can be incorporated without system redesign, (2) \textit{Redundancy}—failure of individual agents doesn't compromise overall system operation, (3) \textit{Distributed Processing}—computational load is distributed across multiple nodes reducing latency, and (4) \textit{Adaptive Learning}—agents can share learned patterns of hazardous situations across different locations. Early deployments suggest that this infrastructure-level approach complements vehicle-based systems, creating a comprehensive safety ecosystem where both autonomous vehicles and infrastructure actively collaborate to prevent crashes before they occur.

\section{Future Directions}
\label{sec:future_directions}

Key research directions for LLM-based crash detection (2025-2026):

\textbf{Efficient Edge Deployment:} Fine-tune LLaVA-1.5 via QLoRA (4-bit, rank $r=16$) targeting AUC $>$88\%, latency $<$100ms on Jetson, memory $<$4GB using sparse attention and TensorRT \cite{hu2021lora}.

\textbf{Causal Reasoning Benchmarks:} Develop CrashCausal-10K with 50K QA triplets (temporal, causal, counterfactual questions) and causal chain annotations linking 4-6 precursor events to outcomes \cite{zarza2023llm, bao2019cadp}.

\textbf{Cross-Geographic Generalization:} Multi-regional datasets (5 continents, diverse vehicle types) with fairness metrics requiring AUC parity within 5\% across regions. Few-shot adaptation (100 examples) targeting $<$4\% drop \cite{sanchez2024equity}.

\textbf{Synthetic Data:} CARLA/SUMO-generated scenarios with hybrid training for rare crashes \cite{dosovitskiy2017carla, fan2024learning}.

\textbf{Emerging Paradigms:} Memory-driven zero-shot detection maintains prototypical ``normal" traffic patterns for deviation-based alerts without crash-specific training \cite{wang2025memoryvad}. Training-free approaches leverage foundation models via prompt engineering for rapid deployment \cite{liu2025trainingfree, zanella2024harnessing}. These paradigms complement fine-tuned models, enabling adaptive systems that switch modes based on data availability and constraints.

\section{Conclusion}
\label{sec:conclusion}

This survey examined LLM/VLM-based video crash detection through early 2025, highlighting both the rapid advancement of the field and the persistent gap between promising research results and real-world deployment constraints. Recent systems, including 2025 innovations like SeeUnsafe for proactive detection and AnomalyCLIP for weakly-supervised recognition, demonstrate strong capabilities in multimodal understanding. However, many works still underreport latency, struggle with cross-domain robustness, and rely on datasets that are relatively small or weakly annotated for safety-critical use. The introduction of new datasets like TU-DAT  (2025) with roadside perspective and fine-grained temporal annotations represents progress, though comprehensive causal chain annotations remain limited. See, e.g., LLM/VLM methods such as Video-LLaMA for video understanding, training-free approaches like LAVAD, and emerging memory-driven architectures for anomaly detection~\cite{zhang2023video,zanella2024harnessing,wang2025memoryvad}.

\subsection{Key Findings}
\begin{enumerate}
\item \textbf{Deployment Gap}: High-capacity VLMs can achieve strong accuracy on curated benchmarks, but real-time, edge-friendly latency is rarely reported and remains a major barrier for on-device ITS deployment. Training-free pipelines (e.g., LAVAD) reduce data/engineering overhead but depend on captioning backends whose speed varies across hardware and models~\cite{zhang2023video,zanella2024harnessing}.

\item \textbf{Generalization Challenge}: Across the broader video anomaly literature, models often degrade when moved off their source domain/dataset; works frequently note domain specificity and the need for adaptation when transferring between datasets such as UCF-Crime, CADP, and DAD~\cite{sultani2018real,bao2019cadp,chan2016dashcam}.

\item \textbf{Data Bottlenecks}: Several widely used datasets for crash/anomaly detection remain modest in size. For example, DAD contains \(\sim\)620 dashcam videos (with positive/negative splits), and UCF-Crime has \(\sim\)1{,}900 untrimmed surveillance videos across 13 anomaly classes~\cite{chan2016dashcam,sultani2018real}. The 2025 introduction of TU-DAT with over 5,000 roadside camera sequences represents progress, though the field still requires larger-scale datasets with causal annotations~\cite{tudat2025}.
\end{enumerate}

\subsection{Critical Recommendations}
\textbf{For Researchers:} Design efficiency-first architectures and report latency alongside accuracy; incorporate domain adaptation/evaluation across datasets; and consider training-free or weakly-supervised regimes where appropriate. Privacy-preserving and federated paradigms are increasingly relevant for multi-site video data~\cite{liu2025privacypreservingvideoanomalydetection}.

\textbf{For Practitioners:} Start with lightweight or training-free baselines (e.g., LAVAD-style pipelines) before committing to heavy retraining; adopt hierarchical processing (quick filters \(\to\) deeper reasoning on suspected segments); plan for local adaptation and monitoring; and keep human oversight in the loop for safety-critical decisions~\cite{zanella2024harnessing}.

\subsection{Synthesis}
We observe: (i) \textbf{Reasoning vs. Deployability}: richer temporal/contextual reasoning often coincides with higher compute cost; (ii) \textbf{Source--Target Drift}: models tuned to one dataset typically require adaptation to new camera placements, traffic patterns, and artifact distributions; (iii) \textbf{System Design Tiers}: in practice, teams converge on edge-first filters with selective offloading to stronger backends when latency budgets permit; (iv) \textbf{Emerging Paradigm Shift (2025)}: The field is witnessing a diversification of approaches—proactive detection systems (SeeUnsafe), weakly-supervised methods (AnomalyCLIP), memory-driven architectures, and training-free frameworks—each addressing different deployment constraints while collectively pushing toward more adaptive and practical systems~\cite{Zhang2025SeeUnsafe,zhou2025anomalyclip,wang2025memoryvad}. These are qualitative patterns—reporting and standardized latency benchmarks remain an open need~\cite{zhang2023video,liu2025privacypreservingvideoanomalydetection}.

\subsection{The Path Forward}
Progress hinges on (a) stronger temporal modeling under efficiency constraints (e.g., video transformers like TimeSformer/ViViT), (b) robust multilingual and low-resource operation via VLM alignment/fine-tuning, (c) privacy-aware training via federated or distributed learning where data cannot leave premises, (d) human-in-the-loop active learning for targeted data curation and assurance, and (e) \textbf{integration of 2025 paradigms}—combining proactive monitoring (SeeUnsafe-style infrastructure integration), weakly-supervised learning (AnomalyCLIP's zero-shot transfer), memory-driven architectures for data-efficient deployment, and training-free approaches for rapid adaptation~\cite{zhang2023video,liu2025privacypreservingvideoanomalydetection,Zhang2025SeeUnsafe,zhou2025anomalyclip}. The convergence of these diverse approaches with enhanced datasets like TU-DAT offers promising pathways toward deployable, adaptive crash detection systems that balance accuracy, efficiency, and real-world practicality~\cite{tudat2025}.

\bibliographystyle{IEEEtran}
\bibliography{main}

\end{document}